\documentclass[letterpaper, 10 pt, conference]{ieeeconf} 
\IEEEoverridecommandlockouts	
\usepackage[dvipsnames]{xcolor}

\usepackage{amssymb}
\usepackage{amsmath}
\usepackage{commath}
\usepackage{mathtools}
\usepackage{import}
\usepackage{paralist}
\usepackage{svg}
\usepackage{textcomp}
\usepackage{gensymb}
\usepackage{dsfont}
\usepackage{subfig}
\usepackage{graphics}
\usepackage{epsfig} 
\usepackage[normalem]{ulem}
\usepackage[binary-units]{siunitx}
\usepackage{bm}
\usepackage{upgreek}
\usepackage{todonotes}
\usepackage{scicite}
\usepackage{times}
\usepackage{xurl}
\usepackage{multirow}
\usepackage{makecell}
\usepackage{tabularx}
\usepackage{booktabs}
\usepackage[percent]{overpic}
\usepackage{transparent}
\usepackage{tikzscale}
\usepackage[htt]{hyphenat}
\usepackage{nameref}
\usepackage{chngcntr}
\usepackage{hyperref}
\usepackage{placeins}
\usepackage{spverbatim}
\usepackage{listings}

\definecolor{codegreen}{rgb}{0,0.6,0}
\definecolor{codegray}{rgb}{0.4,0.4,0.4}
\definecolor{codepurple}{rgb}{0.5,0,0.9}
\definecolor{backcolour}{rgb}{0.95,0.95,0.95}

\lstdefinestyle{mystyle}{
    backgroundcolor=\color{backcolour},   
    commentstyle=\color{codegreen},
    keywordstyle=\color{magenta},
    numberstyle=\tiny\color{codegray},
    stringstyle=\color{codepurple},
    basicstyle=\fontsize{10}{10}\selectfont\ttfamily\ttfamily,
    breakatwhitespace=false,         
    breaklines=true,
    breakindent=0pt,
    captionpos=b,                    
    keepspaces=true,                 
    numbers=none,                    
    numbersep=5pt,                  
    showspaces=false,                
    showstringspaces=false,
    showtabs=false,                  
    tabsize=3
}

\makeatletter
\let\NAT@parse\undefined
\makeatother
\usepackage[numbers,sort&compress]{natbib}

\usepackage{subfig}
\usepackage{hyperref}

\usepackage{xspace}
\usepackage{placeins}

\usepackage[labelfont=bf]{caption}
\usepackage{multicol}
\usepackage{graphicx}
\usepackage{algorithm}
\usepackage{outlines}
\usepackage{booktabs}
\usepackage{bbm}
\usepackage{wrapfig}
\usepackage[noend]{algpseudocode}
\algnewcommand{\IfThenElse}[3]{
  \State \algorithmicif\ #1\ \algorithmicthen\ #2\ \algorithmicelse\ #3}
\algnewcommand{\IfThen}[2]{
  \State \algorithmicif\ #1\ \algorithmicthen\ #2}
\algrenewcommand\algorithmicrequire{\textbf{Input:}}
\algrenewcommand\algorithmicensure{\textbf{Output:}}

\title{RoboCulture: A Robotics Platform for Automated Biological Experimentation}
\author{Kevin Angers\textit{$^{1 *}$}, Kourosh Darvish\textit{$^{1, 2, 4}$}, Naruki Yoshikawa\textit{$^{1, 2}$}, Sargol Okhovatian\textit{$^{1, 3}$}, Dawn Bannerman\textit{$^{1, 3}$},\\ Ilya Yakavets\textit{$^{1, 4}$}, Florian Shkurti\textit{$^{1, 2, 4* \dag}$}, Al\'{a}n Aspuru-Guzik\textit{$^{1, 2, 4, 5, 6*}$}, Milica Radisic\textit{$^{1, 3,4*}$}}

\begin{document}

\twocolumn[{%
\renewcommand\twocolumn[1][]{#1}%
\begin{center}
    \centering
    \maketitle
    \includegraphics[width=0.99\textwidth]{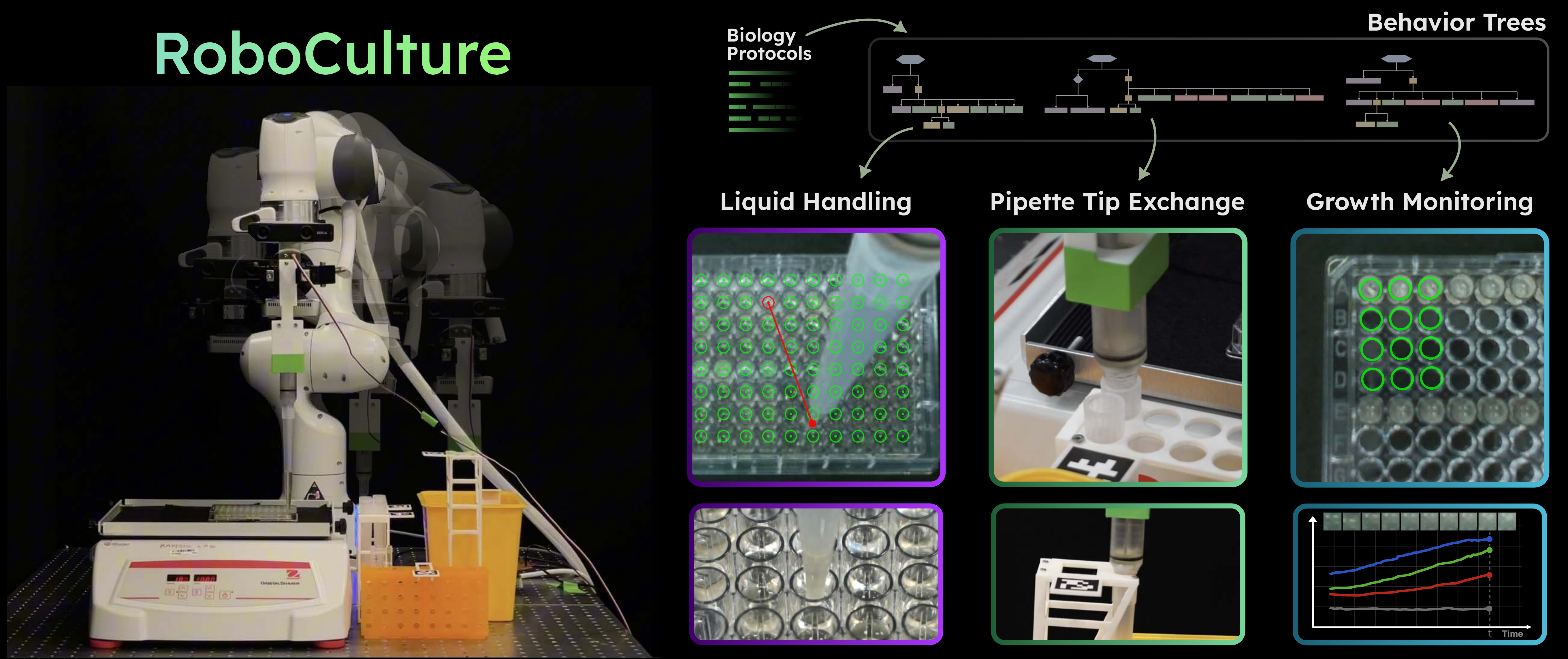}
    \captionof{figure}{RoboCulture integrates a vision‐based liquid handling system capable of reliable pipetting into 96-well plates, a force-guided pipette tip exchange system, and cellular growth monitoring toward generalizable biology laboratory automation. Biology protocols are represented as behavior trees, a reactive and modular framework for experiment state handling. Code, CAD models and video demonstrations of RoboCulture can be found at \href{https://ac-rad.github.io/roboculture/}{https://ac-rad.github.io/roboculture}.}
    \label{fig:fig1}
\end{center}
}]

\footnotetext[1]{University of Toronto, Toronto, ON, Canada}
\footnotetext[2]{Vector Institute, Toronto, ON, Canada}
\footnotetext[3]{Toronto General Health Research Institute, Toronto, ON, Canada}
\footnotetext[4]{Acceleration Consortium, Toronto, ON, Canada}
\footnotetext[5]{Canadian Institute for Advanced Research, Toronto, ON, Canada}
\footnotetext[6]{NVIDIA, Toronto, ON, Canada}
\renewcommand{\thefootnote}{\fnsymbol{footnote}}
\footnotetext[1]{\hangindent=1.8em\hangafter=1 Corresponding authors; E-mails: 
\mbox{kevin.angers@mail.utoronto.ca}, 
\mbox{florian@cs.toronto.edu}, 
\mbox{alan@aspuru.com}, 
\mbox{m.radisic@utoronto.ca}}
\footnotetext[2]{Lead Contact; Email: florian@cs.toronto.edu}

\section*{Summary}

Automating biological experimentation remains challenging due to the need for millimeter-scale precision, long and multi-step experiments, and the dynamic nature of living systems. Current liquid handlers only partially automate workflows, requiring human intervention for plate loading, tip replacement, and calibration. Industrial solutions offer more automation but are costly and lack the flexibility needed in research settings. Meanwhile, research in autonomous robotics has yet to bridge the gap for long-duration, failure-sensitive biological experiments.
We introduce \textbf{RoboCulture}, a cost-effective and flexible platform that uses a general-purpose robotic manipulator to automate key biological tasks. RoboCulture performs liquid handling, interacts with lab equipment, and leverages computer vision for real-time decisions using optical density-based growth monitoring. We demonstrate a fully autonomous 15-hour yeast culture experiment where RoboCulture uses vision and force feedback and a modular behavior tree framework to robustly execute, monitor, and manage experiments.

\paragraph*{Keywords} Flexible Laboratory Automation, Robotics, Cell Culture, Computer Vision, Self-Driving Labs.

\section{Introduction}
\label{sec:introduction}

Biological experimentation is a tedious and time-consuming process. For instance, cell culture often requires careful attention and repetitive manual processes, often necessitating daily maintenance spanning several weeks. The need for manual sub-culturing and culture observation for decision making poses strict requirements on personnel lab attendance that often extend outside regular working hours, thus posing challenges for safety and accuracy in addition to limiting convenience. Naturally, various approaches for automation are often entertained by most labs to minimize these arduous tasks, however, only parts of the overall pipeline have been successfully automated.

Self-Driving Labs (SDLs) built with general purpose robotic manipulators can facilitate complete lab automation, offering the potential for unparalleled flexibility with a relatively small amount of hardware \cite{tom2024self, hase2019nextgenlabs}. SDLs leverage artificial intelligence (AI) and closed-loop experimentation to autonomously design, execute, and analyze experiments, significantly accelerating the rate of discovery by minimizing the need for human intervention. In the biological sciences, SDLs hold particular promise by addressing the variability and complexity inherent in common protocols, providing a pathway toward scalable and reproducible operator-free experimentation across a dynamic range of applications.

Although SDLs represent a comprehensive, long-term vision for laboratory automation, many commercial solutions take a narrower approach by largely focusing on liquid handling, offering an immediate step toward reducing manual workloads. Devices such as the Opentrons OT-2 \cite{opentrons_ot2}, Hamilton Star V \cite{hamilton_star_v} and Flow Robotics Flowbot One \cite{flowbot_one} offer affordable automated pipetting through gantry-based systems designed for high-throughput and precision. 
However, these devices are not autonomous systems; they still rely heavily on human intervention for tasks such as loading and unloading well plates, retrieving reagents from storage, programming assay steps, monitoring cell growth dynamics, and deciding when to sub-culture. 
Consequently, they are not comprehensive solutions for end-to-end laboratory automation, as they address only specific stages of the workflow and continue to depend on human oversight for key experimental decisions and logistical tasks.
Furthermore, automating more complex procedures such as 3D cell culture, including organoid culture, requires more sophisticated pipetting capabilities which are difficult to achieve with standard gantry-based systems. Most commercial liquid handling devices can only pipette perpendicular to the well plate, making them unsuitable for dense cell suspensions and viscous hydrogels, as they cannot effectively prevent bubble entrapment in viscous fluids. Additionally, the delicate and spatially constrained nature of 3D tissues demands pipettes with greater motion flexibility to minimize mechanical stress during pipetting.

Larger industrial systems such as StemCellFactory \cite{elanzew2020stemcellfactory} and Hamilton Cell Care STAR \cite{hamilton_cell_care_star} are designed for ultra-high throughput and end-to-end automation in clinical and industrial settings. However, they are often prohibitively expensive for most laboratories and highly specialized for specific tasks, limiting their adaptability to various workflows. While such systems excel in industrial and clinical settings where large-scale, repetitive processes benefit from their high throughput capabilities, academic laboratories typically prioritize flexibility and the ability to automate a wide variety of tasks over sheer processing speed. Research environments frequently employ a large variety of protocols, making it difficult to integrate rigid, high-cost automation platforms designed for narrowly defined workflows \cite{holland2020lifescienceautomation}. 

Consequently, there is a growing need for flexible, modular platforms that offer generalizable autonomy, as emphasized in recent perspectives on laboratory automation \cite{cooper2025accelerating, angelopoulos2024transforming}. Robotic manipulators have recently been explored for liquid handling, aiming to replicate human-like dexterity and offering greater versatility than conventional gantry-based systems \cite{zhang2022integratingmanualpipettecollaborative, cellcultureautomation_ais}. Despite this, many manipulator-based approaches still rely on extensive manual calibration or task-specific hardware, constraining their capacity to handle diverse tasks and preventing full autonomy across the diverse operations involved in cell culturing.
Often, these systems are constrained to fixed, preconfigured positions for well plates and other labware, where open-loop control is sufficient under ideal conditions but fails to accommodate deviations from the expected setup. Additionally, most existing laboratory automation systems lack closed-loop feedback and real-time perception, leaving them ill-equipped to manage the precision and variability inherent to biological experiments. This gap indicates a critical need for systems that can dynamically adapt to evolving workflows, which remains a major hurdle in realizing next-generation laboratory automation. A more detailed review of related work, including comparisons with recent robotic liquid handling systems, is provided in Section \ref{sec:related-work}.
Furthermore, while recent advancements in robotics learning, for example, those leveraging reinforcement learning and foundation models~\cite{firoozi2023foundation}, have demonstrated impressive capabilities in various environments, their application in real-world laboratory settings remains limited. These models currently lack the robustness needed for long-horizon, failure-intolerant biological protocols, and their training on broad, non-specialized datasets poses challenges when adapting to the nuanced demands of biological research.

Standard cell culture is typically conducted in multi-well plates (127.76 × 85.48 $\text{mm}^2$) by combining cells with supportive culture media. An experienced scientist seamlessly integrates sequential pipetting of cells and culture media into designated wells at controlled volumes, at speeds that ensure maintenance of cell viability. The human operator will then monitor cell culture growth using a microscope or a plate reader at timed intervals to gather information about cell expansion to determine the appropriate time for sub-culturing, i.e. removing the concentrated cell suspension for further expansion or product collection. These decisions are informed by ongoing observations and often require flexibility, as the timing of cell growth can be highly variable and difficult to predict. This inherent stochasticity of living systems makes it challenging to pre-program such decisions into a robotic system, posing a significant barrier to full automation.

\begin{figure*}[h]
    \centering
    \includegraphics[width=0.99\linewidth]{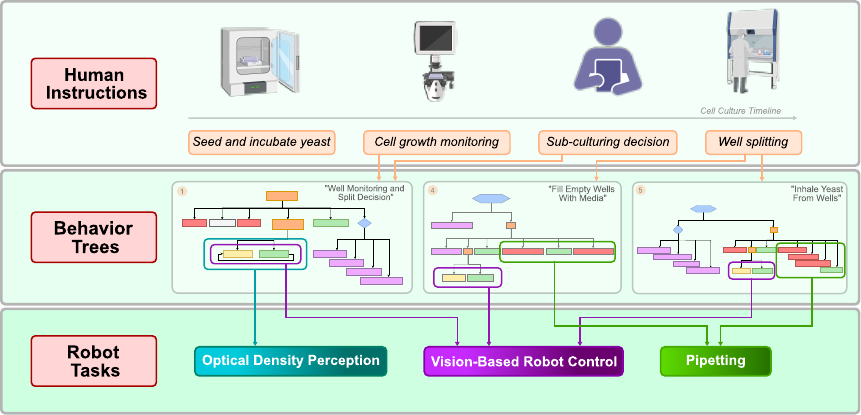}
    \caption{RoboCulture bridges human-level biological protocols and low-level robotic tasks using behavior trees. High-level experimental instructions such as growth monitoring and sub-culturing yeast are translated into modular, hierarchical behavior trees composed of individual behaviors. These trees coordinate key robotic subsystems, including optical density perception, vision-based robot control, and pipetting, enabling autonomous execution and decision making during complex cell culture workflows. A more detailed description of the behavior trees and individual behaviors is provided in Section \ref{sec:behavior-trees}.}
    \label{fig:bio-to-robo}
\end{figure*}
    
\begin{figure*}[t]
    \centering
    \includegraphics[width=1.0\linewidth]{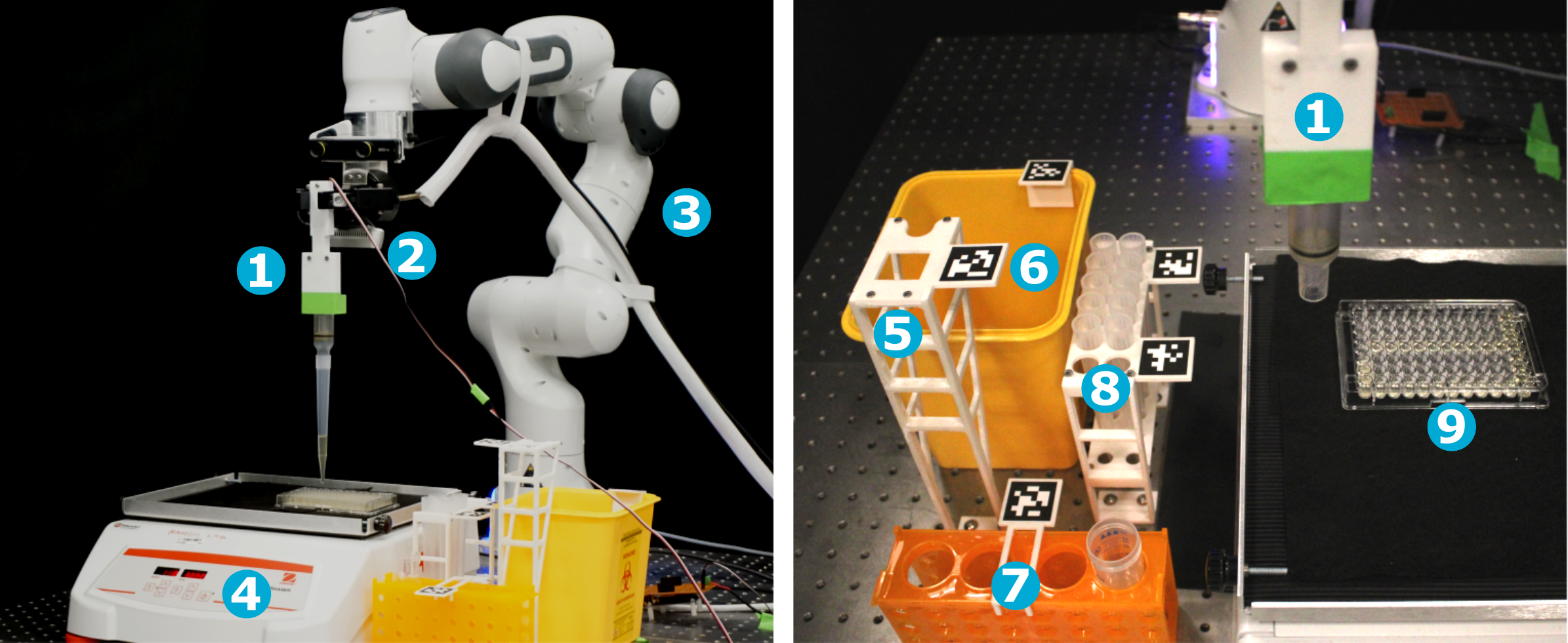}
    \caption{The yeast growth experiment setup. 1) Our Digital Pipette v2, 2) a downwards-facing Intel RealSense D435i camera, 3) a Franka Emika robot with a Robotiq 2F-85 gripper, 4) an OHAUS SHHD1619DG Heavy Duty Orbital Shaker Platform, 5) a 3D printed pipette tip remover, 6) a biological waste bin, 7) a Falcon Tube rack holding YPD media, 8) a 3D printed pipette tip rack, and 9) a 96 well plate prepared with yeast.}
    \label{fig:setup}
\end{figure*}

\textbf{Contributions.}
Here, we introduce \textbf{RoboCulture}, a step towards end-to-end autonomy of biological experimentation with general purpose robotics (\textbf{Figure \ref{fig:fig1}}). It is capable of reliably performing liquid handling-based assays as well as other related tasks which leverage the capabilities of a robot manipulator, including pipette tip exchange and growth monitoring, to fully assist scientists with end-to-end cell culture operations. RoboCulture presents the following unique advantages:
\begin{enumerate}
    \item It allows for the automation of liquid handling experiments with a general purpose 7-axis robotic manipulator. RoboCulture comprises three main components: a) a computer vision-based liquid handling system which uses vision feedback to establish closed-loop control and robust insertion of the pipette into wells of an arbitrarily positioned 96-well plate, b) a pipette tip exchange system leveraging robot force feedback to reliably attach new pipette tips, and c) a computer vision system to monitor the optical density of the wells to inform RoboCulture on the current state of the experiment (\textbf{Figure \ref{fig:fig1}}).

    \item RoboCulture leverages our Digital Pipette v2, an updated version of our open-source, cost-effective Digital Pipette \cite{digital-pipette}. This iteration features interchangeable pipette tips, making it suitable for use in biomedical settings that require sterility. The pipette is designed for interoperability with a robot gripper without requiring extensive hardware modifications for compatibility. We release the code and CAD models which are adaptable and customizable for new tasks.
    
    \item RoboCulture autonomously performs a 15-hour yeast cell culture experiment, integrating key tasks such as liquid handling, pipette tip exchange, growth monitoring, and decision-making for sub-culturing. A behavior tree governs the coordination between these subsystems, as illustrated in \textbf{Figure~\ref{fig:bio-to-robo}}, while \textbf{Figure~\ref{fig:setup}} depicts the overall experimental setup.
\end{enumerate}

RoboCulture is designed to prioritize flexibility and autonomy over throughput. It enables long-duration, hands-free operation by responding to changing experimental conditions. Its modular behaviors can be reused and combined to define new protocols, making it well-suited for research settings where adaptability is more important than speed.

We release the code and CAD models which are adaptable and customizable for new tasks.
\section{Results}
\label{sec:results}

To enable autonomous cell culture, we began by translating the sequence of human-performed tasks into operations that could be executed by a robot. In a standard workflow, a human operator typically seeds cells into a well plate, places the plate on a shaking platform and periodically monitors cell growth by transferring the plate to a microscope or plate reader. Based on these observations, the operator performs offline analyses to assess cell density and decides whether sub-culturing is required. If so, the contents of a saturated well are split into fresh wells to promote continued proliferation. 
To implement these procedures autonomously, RoboCulture uses behavior trees \cite{colledanchisebtrees} to translate human-level protocols into low-level robot tasks (\textbf{Figures \ref{fig:btree-main}-\ref{fig:btree-remove-tip}}). We design a set of modular behaviors that encapsulate key robotic capabilities including optical density perception, vision-based robot control, and pipetting. These behaviors are composed into a hierarchical structure that allows RoboCulture to execute complex workflows and make decisions about when to initiate well splitting (\textbf{Figure \ref{fig:bio-to-robo}}). The initial seeding of cells remains a manual step performed prior to the start of the autonomous experiment.

\subsection{RoboCulture Design}

RoboCulture is built around a 7-axis Franka manipulator equipped with an RGB-D camera and standard laboratory equipment, including a digitally controlled orbital shaker to keep cells in suspension, well plates, a waste bin, and racks for pipette tips and conical tubes (\textbf{Figure \ref{fig:setup}, Figures \ref{fig:tiprack}-\ref{fig:tipremover}}). We use AprilTags~\cite{olson2011tags} to localize these static components, which are fiducial markers commonly used in robotics to denote the positions of objects. These tags must be positioned such that each tag is visible from a central ``home" position beneath the downward-facing camera. Whenever the layout of the workspace changes, a calibration procedure is performed to establish pixel offsets between each AprilTag and the desired end-effector pose for a given object, allowing the robot to accurately detect and move between the pieces of lab hardware. While fiducial markers are sufficient for static equipment that does not require high precision, they are unsuitable for well plates, which demand far greater accuracy; thus, no AprilTags are placed on the well plate.

We first optimized individual robotic tasks such as pipetting, vision-based robot control, and optical density perception, and then demonstrated the effectiveness of RoboCulture by autonomously performing a yeast culturing experiment, which lasted 15 hours, on a 96-well plate.

\subsection{Pipetting Tasks}
\subsubsection{Accuracy of Liquid Dispensing}
Accurate pipetting is essential for effective cell culture procedures. We present Digital Pipette v2, an upgraded version of our cost-effective Digital Pipette originally introduced in \cite{digital-pipette}, redesigned for biomedical applications. A key improvement is the replacement of the fixed syringe with removable 10 mL pipette tips, in accordance with standard practices that require disposable tips to prevent cross-contamination. To accommodate this change, we modified the pipette design from a positive (direct) displacement (Type D) pipette to an air displacement (Type A) pipette \cite{ISO8655-2_2022}. Since sterility must be maintained throughout the procedure, tips must be exchanged between each pipetting operation. To facilitate this, we developed a 3D-printed mechanism capable of reliable tip exchange. 

We evaluated the pipette's liquid dispensing accuracy using a gravimetric testing procedure described by the International Standard on Piston-Operated Volumetric Apparatus (ISO 8655-6) \cite{ISO8655-6_2022}. In this procedure, the weight of dispensed volume is measured with a balance, and the systematic and random error are computed from the mean delivered volume over multiple iterations at target volumes of 1 mL, 5 mL and 10 mL. \textbf{Table \ref{tab:gravimetric_test}} summarizes the results, reporting the mean delivered volume $\bar{V}$, the standard error $\mu_s$, and the random error $C_v$. The errors observed were significantly below the maximum permissible limits defined by ISO 8655-2 \cite{ISO8655-2_2022}, indicating very high repeatability of the Digital Pipette v2.

\begin{table}[ht]
\centering 
\begin{tabular}{lllll}
\toprule
Volume (mL) & Device/Standard & $\overline{V}$ (mL) & $\eta_s$ (\%)        & $C_v$ (\%)       \\ \hline
10.0        & Digital Pipette v2 & 9.9909 & \textbf{-0.09} & \textbf{0.10} \\
            & ISO 8655        & -      & 0.6            & 0.3           \\ \hline
5.0         & Digital Pipette v2 & 4.9949 & \textbf{-0.10} & \textbf{0.16} \\
            & ISO 8655        & -      & 1.2            & 0.6           \\ \hline
1.0         & Digital Pipette v2 & 0.9951 & \textbf{-0.49} & \textbf{0.58} \\
            & ISO 8655        & -      & 6.0            & 3.0
        \\
\bottomrule            
\end{tabular}
\caption{Digital Pipette v2 Gravimetric Test. Comparison of the standard error $\eta_s$ and the random error $C_v$ obtained from the gravimetric test with the maximum permissible errors defined by ISO 8655-6. }
\label{tab:gravimetric_test}
\end{table}

\begin{figure}
    \centering
    \includegraphics[width=1.0\linewidth]{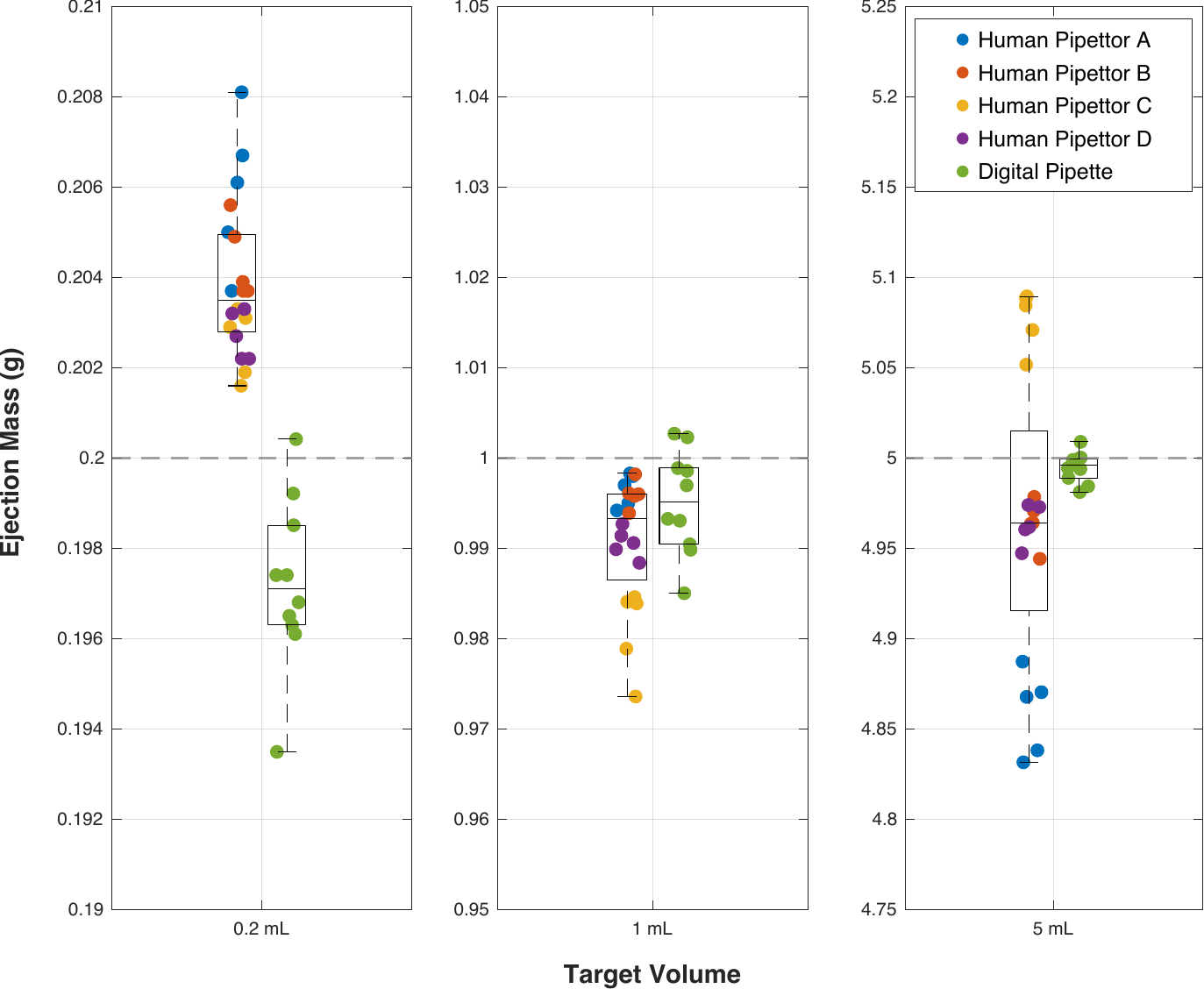}
    \caption{Gravimetric comparison of the Digital Pipette v2 with human pipettors, for volumes of 0.2 mL, 1 mL and 5 mL. A Levene’s test was conducted to compare the variances between the two groups (Digital Pipette v2 vs. Human).}
    \label{fig:robot-vs-humans}
\end{figure}

In addition, four experienced human operators performed the same gravimetric testing procedure to compare the reliability of the Digital Pipette v2 with that of manual pipetting (\textbf{Figure \ref{fig:robot-vs-humans}}). To account for the small volumes typically used in yeast cell culture, we tested dispensing at 0.2 mL, 1 mL, and 5 mL. Each operator performed five replicates at each volume. For the 0.2 mL and 1 mL volumes, a standard P1000 pipette was used as the human baseline, while a 5 mL serological pipette was used for the 5 mL volume. Operators were instructed to carefully aspirate the desired volume and dispense it in full, mirroring the procedure followed by the Digital Pipette v2. For the 0.2 mL and 1 mL tests, there were no significant differences in variance between the two groups (p = 0.8754 and p = 0.6533, respectively). However, for the 5 mL test, the Digital Pipette v2 demonstrated significantly lower variance (p = 0.0046).

\subsubsection{Vision-Based Pipette Tip Positioning}
Having validated that the Digital Pipette v2 performs equal to or better than experienced human operators in volume dispensing accuracy, we next focused on refining pipette insertion and tip exchange tasks. Precise positioning of the pipette tip is critical for RoboCulture's liquid handling system, but several sources of error make accurate insertion into the 9 mm-diameter wells of a 96-well plate challenging. The most significant of these errors stem from RGB camera calibration with respect to the base of the robot. Fiducial markers are commonly used in computer vision to convert an object’s position in a camera image into its world pose, its position and orientation relative to the robot’s base. This conversion is dependent on camera extrinsic calibration, which is very sensitive to small errors that could lead to significant positioning errors–see Section \ref{sec:perception-error} for a more detailed justification on camera calibration error.
Another major source of error is the uncertainty in the pipette tip’s location. While the robot knows the pose of its end-effector through internal forward kinematics, it does not have direct knowledge of the pipette tip’s true pose in 3D space. This is because the gripper does not consistently grasp the pipette at precisely the same location, and even small variations at the grasping point can result in substantial deviations at the tip. The transformation between the robot base and the end effector, $T_{ee}^0$ , is known, but the transformation between the end effector and the pipette tip, is not. Attempting to approximate or hard-code this transformation introduces uncertainty, as grasping errors can cause the true pipette tip position to deviate from expectations.

For this experiment, we placed the 96-well plate at random locations, and instructed the robot to insert the pipette tip into each well in a randomized order, achieving a successful insertion rate of 99\% (\textbf{Table \ref{tab:pipette-positioning}, Supplementary Video 1}). The overall success rate is presented, where a success is any insertion such that the pipette entered the well, including those which required a ``retry". A retry occurs when the pipette presses down on the edge of the well, and the large increase in the robot end-effector force initiates another insertion attempt, ultimately leading to a successful insertion. Ideally, we would like to minimize cases where a retry is necessary, so we also present the ``perfect" rate; where a perfect insertion is where the pipette perfectly enters the well without touching the edge or requiring any retries. A failed insertion is where the pipette is unable to insert the pipette even after re-attempting the insertion, and are generally caused by perception errors.

\begin{table*}[ht]
    \centering
    \begin{tabular}{lcccccc|c} 
    \toprule
        Trial Number & \textbf{1} & \textbf{2} & \textbf{3} & \textbf{4} & \textbf{5} & \textbf{6} & \textbf{Totals} \\
        \midrule
        Num. Attempts & 96 & 96 & 96 & 96 & 96 & 96 & 576 \\ %
        Num. Retries (Successful insertion after retry)  & 7 & 5 & 7 & 0 & 5 & 14 & 38 \\
        Num. Fails (Unsuccessful insertion)  & 1 & 2 & 3 & 0 & 0 & 0 & 6 \\
        \midrule
        \textbf{Success Rate (Num. insertions without fail / num. attempts)} & \textbf{0.990} & \textbf{0.979} & \textbf{0.969} & \textbf{1.000} & \textbf{1.000} & \textbf{1.000} & \textbf{0.990} \\ 
        Perfect Rate (Num. insertions without fail or retry / num. attempts) & 0.917 & 0.927 & 0.896 & 1.000 & 0.948 & 0.854 & 0.924 \\

        \bottomrule
    \end{tabular}
    \caption{Pipette tip insertion experiment results. For each random position of the well plate, the robot was instructed to insert the pipette into the wells following a randomized sequence of 96 wells. 
    A retry occurs when the robot attempts to insert the pipette but it presses against the edge of the well without entering, but the increase in force triggers another insertion attempt which ends up being successful. A maximum of three retries per well are allowed before it is considered a failed insertion attempt. The number of retries and fails is out of 96 and is used to calculate the success rate and perfect rate, where the success rate allows successful insertions following a retry, but the perfect rate represents only trials succeeding on the first attempt. Six random locations for the well plate were tested.}
    \label{tab:pipette-positioning}
\end{table*}

\begin{figure*}[t]
\centering
\includegraphics[width=0.9\linewidth]{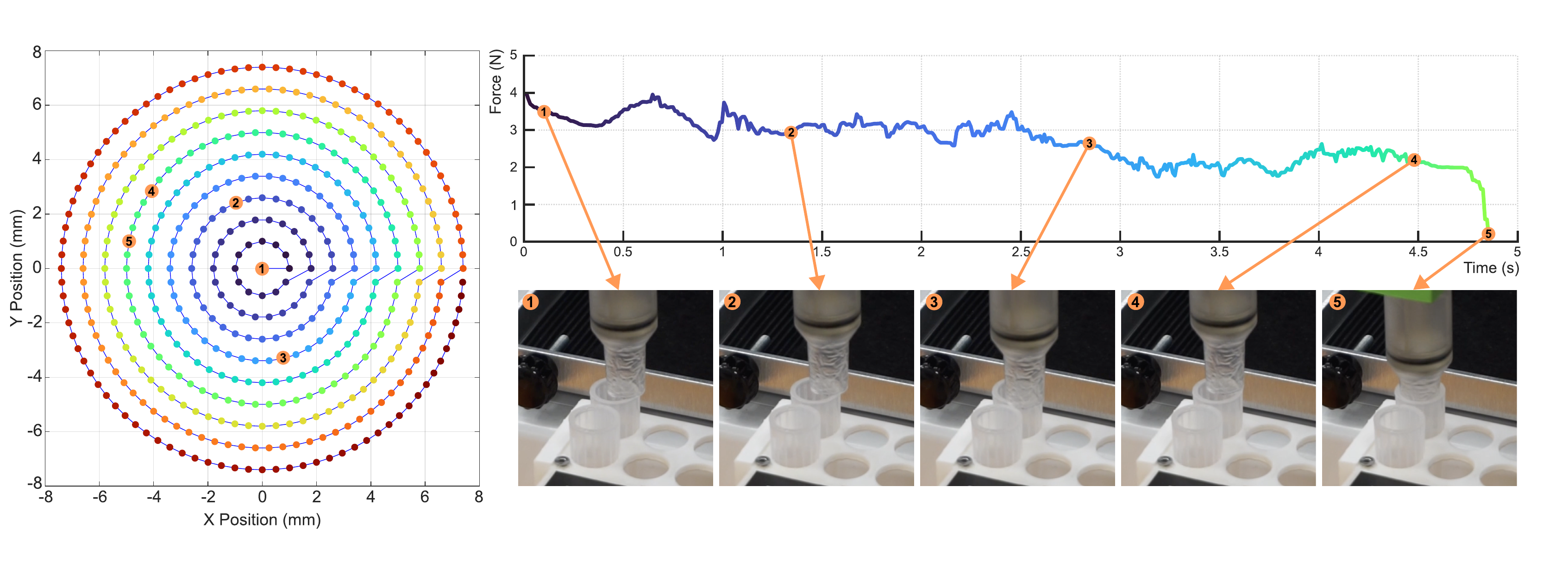}
\caption{Demonstration of the spiral search process for pipette tip attachment. Initially, the pipette body and the new tip are misaligned. A spiral trajectory of waypoints in the XY plane is generated around the start position of the pipette body, and the robot moves the pipette along this trajectory while pressing the pipette onto the surface of the new tip. The force at the end-effector is monitored during the trajectory, and the pipette is finally lowered into the pipette tip when the force drops considerably.}
\label{fig:pipette_force_spiral}
\end{figure*}

\subsubsection{Pipette Tip Exchange}
Pipette tip attachment presents a similar challenge to guiding the pipette tip to a well for liquid handling, but instead of aligning the tip with a well, we must align the pipette body with the opening of a new tip inside a tip rack. The tight insertion required for tip attachment demands extremely precise alignment, making open-loop methods infeasible. Our single camera setup poses additional difficulty for computer vision-based methods, as most of the new tip is occluded by the pipette body. Instead, we leverage the perceived end-effector force to guide the alignment, framing the problem as a peg-in-hole task and employing a spiral search strategy to locate the proper insertion point (\textbf{Figure \ref{fig:pipette_force_spiral}}). 
In a spiral search, the robot moves the pipette body in a controlled spiral trajectory, while monitoring the applied force at the end effector caused by pressing the pipette body onto the edge of the new pipette tip (\textbf{Supplementary Video 2}). The alignment is achieved as follows: First, the robot lowers the pipette body onto the edge of the new tip, applying firm pressure. The initial alignment provided by the AprilTag is expected to not result in a successful insertion, since the tolerance is very small. Second, a spiral trajectory is generated in the XY plane, and the robot moves along this path while pressing the pipette body on the new tip. The force at the end-effector is internally estimated using the robot’s joint torque sensors\footnote{Specifically, the Franka Emika robot estimates Cartesian wrench at the end-effector by transforming joint torques through the Jacobian transpose: $\bm{f}_{ee} = (\bm{J}^T)^{-1} \bm{\tau}_{meas}$, where $\bm{J}$ is the geometric Jacobian at the end-effector and $\bm{\tau}_{meas}$ is the vector of joint torques measured by the robot’s internal torque sensors at each actuator. These sensors detect the effort required to maintain each joint’s position, allowing the robot to infer external forces acting on the end-effector.} and monitored continuously during the spiral trajectory (\textbf{Figure \ref{fig:pipette_force_spiral}}). Third, when a significant drop in force is detected, it indicates that the pipette body has found the correct alignment. The robot then lowers the pipette into the tip to complete the attachment.
For pipette tip removal, the robot guides the pipette toward a 3D-printed tip remover designed to hook the edge of the tip. The robot then lifts upwards, detaching the tip and allowing it to drop into a biomedical waste bin below (\textbf{Supplementary Video 3}).
We performed multiple trials involving sequential tip attachment and removal procedures in order to evaluate the effectiveness of our pipette tip exchange system (\textbf{Supplementary Video 4}). A total of 36 attachment and removal cycles were performed, with 12 cycles executed at each of three distinct pipette grasp locations. The system achieved 100\% accuracy in both attachment and removal tasks, with an average spiral search time of 9.08 seconds per tip insertion, with a standard deviation of 5.76 seconds.

\subsection{Visual Servoing for Well Plate Liquid Handling}

\begin{figure*}[t]
\centering
\includegraphics[width=0.9\linewidth]{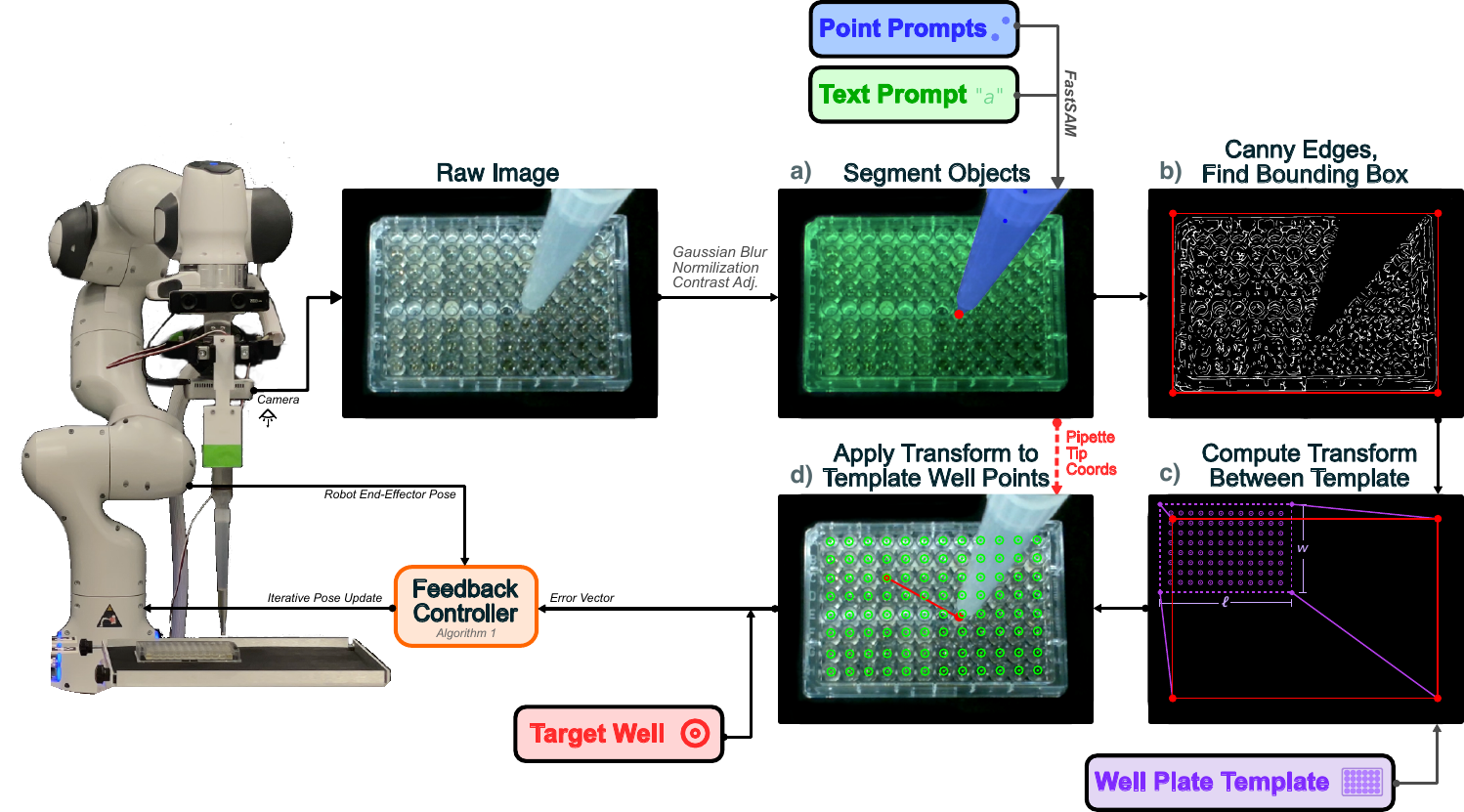}
\caption{Overview of the perception pipeline. Raw images are obtained from the downward facing camera attached to the robot’s end effector. The raw images are preprocessed with a Gaussian blur, normalization and contrast adjustment. The pipette and the well plate are segmented using FastSAM, using point prompts and a text prompt as inputs respectively. The image is converted into a binary edge representation using the Canny algorithm, and masked using the mask obtained from the well plate segmentation. The bounding rectangle produces the corners, which are used to compute a projective transformation to a template well plate. This transform is applied to the template well locations to obtain the image coordinates of each well. The pipette tip coordinates are found using the pipette tip segmentation mask. The error vector between the pipette tip and a desired well can be computed in the image frame and used to control the robot motion via a feedback controller.}
\label{fig:perception}
\end{figure*}

Following the development of accurate volume dispensing, the next challenge was achieving the positional precision required to insert the pipette into the narrow wells of a 96-well plate. To achieve precise and reliable pipetting across 6-, 24-, and 96-well formats, we developed a computer vision pipeline that enables image-based visual servoing–a method first introduced in \cite{sri1979real}. \textbf{Figure \ref{fig:perception}} outlines the perception algorithm used for this task. 

The pipeline extracts key image features from frames captured by a downward-facing camera mounted on the robot, allowing it to localize both the pipette tip and the target well in the image. The robot then minimizes the 2D pixel-space error vector between these two points through closed-loop control.
By controlling the robot using pixel distances rather than estimating well locations in the world frame, we mitigate errors stemming from calibration and improve robustness. Visual servoing is constrained to the XY plane of the end-effector under the assumption of a fixed plate height, while the Z component is independently controlled (\textbf{Supplementary Videos 5, 6}). 

\begin{figure*}[t]
\centering
\includegraphics[width=0.9\linewidth]{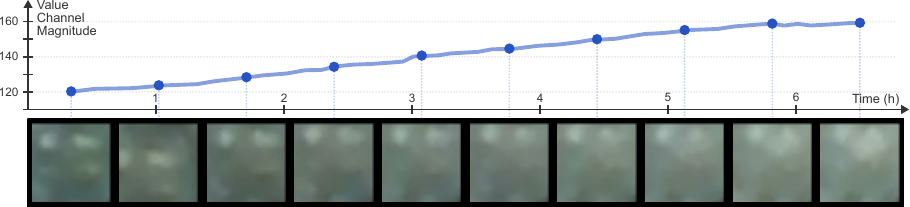}
\caption{Demonstration of well growth monitoring. RoboCulture captures images of the wells over time, extracting the magnitude of the value channel as a measure of image brightness. Changes in optical density are tracked to generate a growth curve.}
\label{fig:growth}
\end{figure*}

\subsection{Optical Density Perception}
Biological laboratories commonly use plate readers to measure a sample's optical density as an indicator of cell confluency \cite{Beal2020, mira2022estimating}. Optical density quantifies how much light at a particular wavelength is absorbed by the sample, and it provides a useful estimate of cell confluency in most cases of suspended cell cultures, since turbidity generally increases as higher cell counts scatter light more effectively. By shining light through the sample and measuring the transmitted intensity, the plate reader data can be calibrated post-hoc against hemocytometer counts to relate optical density values to actual cell density. 
In RoboCulture, we used standard RGB images to offer comparable growth monitoring, avoiding the need for a dedicated plate reader while still enabling informed decisions about the experiment progress leading to faster execution times. Rather than deriving exact cell counts, we focused on tracking relative optical density changes over time (\textbf{Figure \ref{fig:growth}}). It is important to track relative changes in optical density, since estimates using an RGB image are very sensitive to external factors, such as lighting conditions and reflections. We assumed these conditions remain stable so that repeated measurements reliably mirror true growth trends without environmental interference. During experimentation, the robot collects images of the centers of each well at regular intervals to construct a growth curve based on the optical density of the wells (\textbf{Supplementary Video 7}).

\subsection{Behavior Tree Decision Making}
Core cell culture tasks can be replicated by an autonomous robot capable of pipetting, visual servoing and optical density perception, yet the order and timing of these tasks are critical for successful execution. To manage this complexity, we designed a behavior tree that enables RoboCulture to determine which task to execute based on the current state of the experiment (\textbf{Figures \ref{fig:btree-main}-\ref{fig:btree-remove-tip}}). Behavior trees provide a structured framework for modular and reactive decision-making, offering greater adaptability than traditional finite state machines, which often rely on rigid, predefined state transitions. 
Behavior trees organize tasks and conditions into reusable components called ``behaviors", and are arranged hierarchically to reflect task priority. The tree is traversed from the root node downward, ensuring that higher-priority actions override lower-priority ones when necessary. This structured execution enables real-time adaptation without requiring cumbersome, hard-coded state transitions. Because behaviors are independent and easily rearranged, behavior trees simplify the development, debugging, and extension of automation workflows, avoiding the rigidity that typically accompanies complex finite state machines. Detailed descriptions of the behavior trees and behaviors used for RoboCulure are provided in Section \ref{sec:behavior-trees}.

\begin{figure*}[ht]
    \centering
    \includegraphics[width=1.0\linewidth]{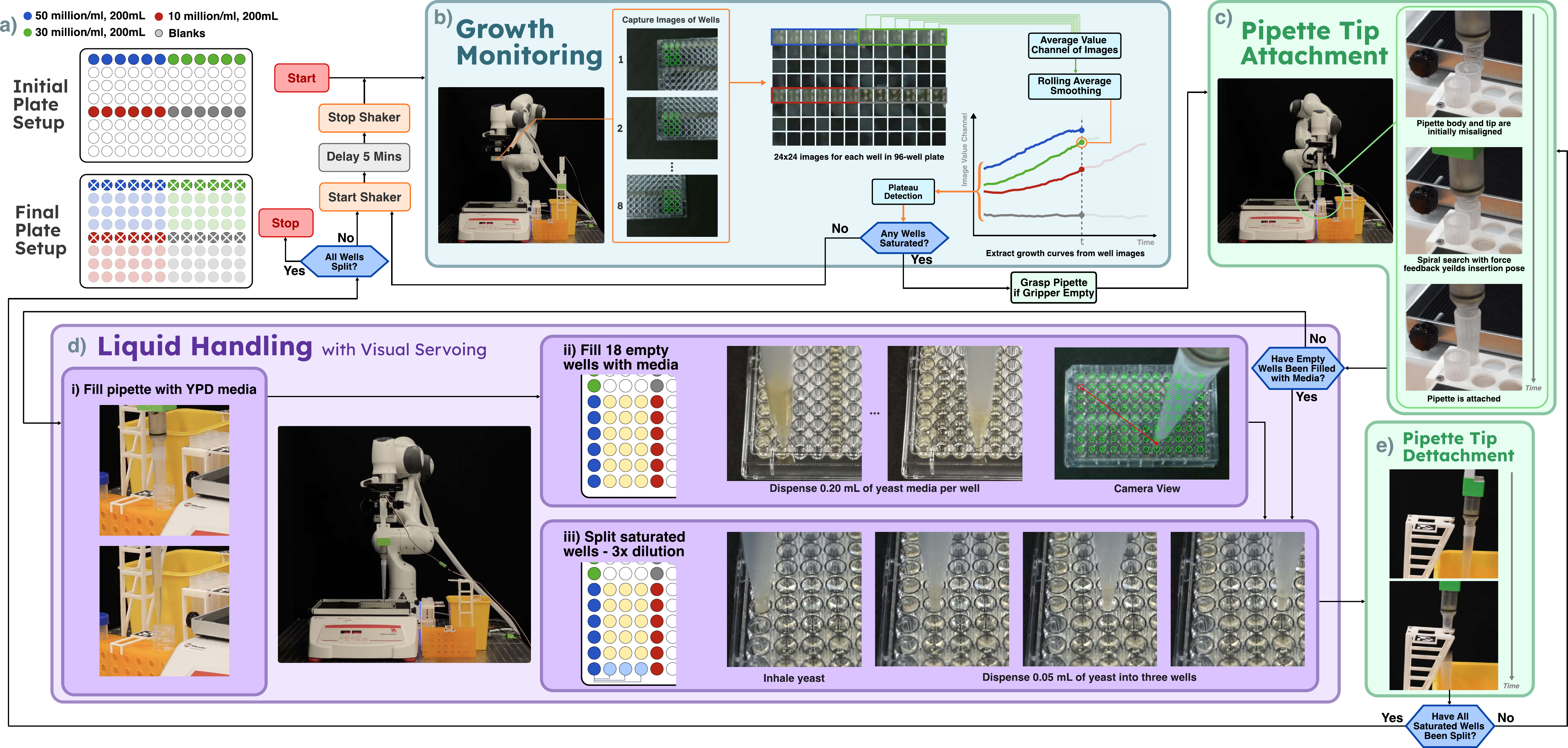}
    \caption{The yeast growth experiment procedure. a) The configurations of the well plate before and after the experiment, showing the different initial concentrations and splitting pattern. Each group is split into the three rows below it on the well plate, and the original wells are voided. b) The growth monitoring stage: images of the wells are taken at regular intervals to extract a growth curve, which is used to inform the progression of the experiment. c) The pipette tip attachment stage: new pipette tips are reliably attached using force feedback. d) The liquid handling stage: i) the pipette is filled with media, ii) a visual servoing based control strategy is used to reliably dispense media into empty wells, and iii) yeast is aspirated from the saturated wells and dispensed into three media-filled wells. e) The pipette tip detachment stage: the contaminated tip is removed from the pipette. CAD models for the Digital Pipette v2, the pipette tip rack and the pipette tip remover are available on our project website.}
    \label{fig:process}
\end{figure*}

\subsection{Autonomous Yeast Culture Experiment}
Finally, the previously described components were integrated into a fully autonomous yeast culturing experiment (\textbf{Figure \ref{fig:process}, Supplementary Videos 8-13}). For this demonstration, we used \textit{Saccharomyces cerevisiae}, a widely studied, non-pathogenic yeast, and conducted all procedures in a Biosafety Level 1 (BSL-1) facility. As the yeast proliferated, the optical density of each well increased until the cell population reached the maximum supported by the available culture media. At this point, a human operator would typically split the saturated well into several fresh wells to enable continued growth. Here, we describe how RoboCulture performed these tasks autonomously.

The goal of the experiment was to autonomously expand yeast cells in suspension and perform a serial dilution. A human operator initially prepared a 96-well plate with six replicates of three different initial concentrations (50 million cells/mL, 30 million cells/mL, 10 million cells/mL) as well as six blank wells of only growth media (\textbf{Figure \ref{fig:process}a}). These conditions were selected to ensure that different wells reached saturation at different times, allowing us to evaluate RoboCulture’s ability to make accurate decisions about when to perform well-splitting. Once the wells have reached the desired density, the saturated wells were split into three new wells, and the previously saturated wells were voided. The blank group was split at the end of the experiment.

To keep the yeast cells in suspension, the well plate was placed on an orbital shaker platform that continuously agitated the samples and prevented the yeast from settling to the bottom of the wells. Every five minutes, RoboCulture stopped the shaker and captured images of each well to assess cell growth, yielding a growth curve representing yeast proliferation in the well (\textbf{Figure \ref{fig:process}b}). If none of the groups reached saturation, the shaker was reactivated for another five minutes, and the monitoring cycle resumed. Once a saturated well was detected, RoboCulture initiated the well-splitting procedure. 

For well-splitting, the robot first grasped the Digital Pipette v2 from its stand if it was not already being held, and moved to the pipette tip rack to attach a new tip (\textbf{Figure \ref{fig:process}c}). RoboCulture then filled the pipette with yeast extract peptone dextrose (YPD) media by lowering it into a Falcon tube and actuating the plunger to draw enough liquid to fill the empty wells. Then, the three empty wells located below each saturated well were filled with 0.2 mL of media (\textbf{Figure \ref{fig:process}d}); these target wells are shown in \textbf{Figure \ref{fig:process}dii}. Immediately afterward, the splitting process began. 
Starting with the leftmost saturated well, RoboCulture first re-suspended each sample by pipetting up and down in a repeated manner to mitigate the yeast settling on the bottom of the plate. Then, the saturated well was inhaled, and 0.05 mL was dispensed into each of the three pre-filled wells below. The original well was voided, and the pipette tip was detached. This procedure was repeated for each of the remaining five saturated wells, with a new tip attached for each split (\textbf{Figure \ref{fig:process}diii}). Once all wells were split, RoboCulture resumed the shaker and re-initiated growth monitoring.

\begin{figure*}
    \centering
    \includegraphics[width=1.0\linewidth]{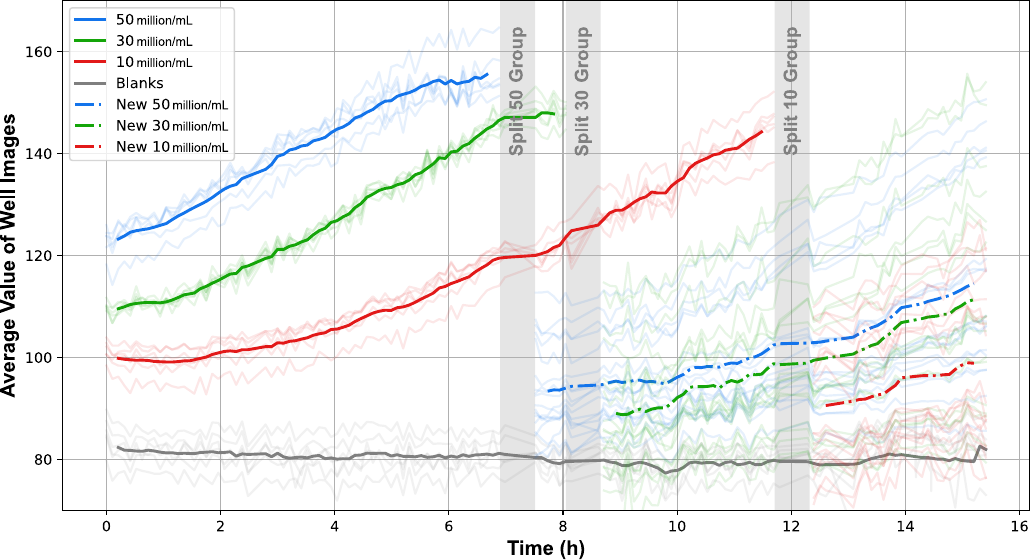}
    \caption{Results from the yeast culture experiment. The y-axis shows image brightness, which correlates with the optical density of each sample. Colored solid lines represent the growth of three experimental groups with initial yeast concentrations of 50 million, 30 million, and 10 million cells/mL, respectively. The solid gray line denotes the blank group. Dashed lines indicate the growth of new wells following 3× dilution. Faint lines correspond to raw image values from individual replicates (n=6) per group, while opaque lines show the rolling average (window size of 5) for each group. Vertical gray bands mark the time points at which splitting occurred. The experiment ran for over 15 hours.}
    \label{fig:yeast-growth}
\end{figure*}

\textbf{Figure \ref{fig:yeast-growth}} shows the results of the yeast culture experiment, illustrating the growth curves for the three experimental groups. The curves follow a typical growth pattern, beginning with an initial lag phase, transitioning into an exponential phase, and eventually reaching a stationary (saturated) phase. As expected, RoboCulture first split the wells seeded with high cell density (50 million cells/mL) at 6.5 hours, followed by the intermediate group (30 million cells/ml) at 8 hours followed by the low density group (10 million cells/mL) at 12 hours. Due to the slower growth in the lowest-density group, these wells had not yet reached saturation. RoboCulture was manually instructed to split these wells pre-emptively due to time constraints. A photo of the final well plate is shown in \textbf{Figure \ref{fig:final-well-plate}}.

Importantly, the newly subcultured wells, indicated by dashed lines (\textbf{Figure \ref{fig:yeast-growth}}), exhibited similar growth patterns as their parent wells, consistent with the high, intermediate and low initial seeding densities. 
The increased variance in the raw curves (faint lines) after the split is primarily due to small inconsistencies in the volume of yeast delivered to each daughter well, which affect the resulting optical density. However, since growth is tracked based on the relative change within each well, the method remains reliable for monitoring proliferation trends.

Furthermore, the exponential phase appeared linear because optical density, rather than cell count, was being measured. At low to moderate concentrations, optical density increases proportionally with cell concentration before deviating at higher densities due to light scattering effects. The blank group did not show growth, suggesting that contamination was unlikely, and further validating the accuracy of our growth monitoring protocol.

\begin{figure}
    \centering
    \includegraphics[width=1.0\linewidth]{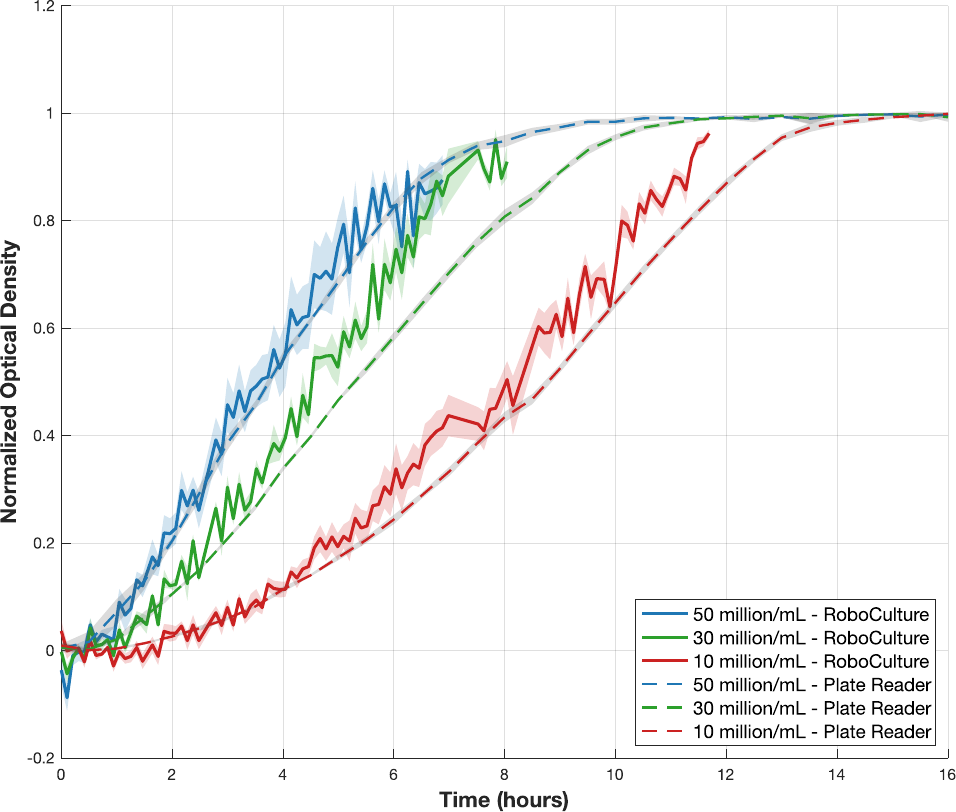}
    \caption{Comparison of plate reader growth curves (dashed) with our growth curves obtained from the robot's camera (solid).}
    \label{fig:rgb_vs_plate_reader}
\end{figure}

To additionally validate the use of the RGB camera for measurement of the optical densities of the wells, we compared the growth curves obtained by the camera during the yeast growth experiment with those obtained using a plate reader from cultures maintained by a skilled human operator. The plate reader measurements were taken from an identical plate prepared simultaneously with the one used in the robotic experiment. The results, shown in \textbf{Figure \ref{fig:rgb_vs_plate_reader}}, are normalized and overlaid due to differences in measurement units. For the camera derived curves, the initial 10 data points were averaged and used as the baseline value during the normalization process. The RoboCulture data showed reasonable agreement with the growth curves obtained from a standard cell culture by a skilled human operator at identical initial seeding concentrations, supporting the effectiveness of the autonomous cell culture protocol.

\section{Discussion}

Building on recent advancements in automated pipetting with general-purpose manipulators, we developed a method that enables reliable pipette insertion into the wells of a 96-well plate without requiring prior calibration or knowledge of its position. Given that each well measures only 9 millimeters in diameter, precise positioning is crucial to avoid damaging samples or spilling hazardous biological material. To address this, we implemented an image-based visual servoing controller that dynamically compensates for both perception and mechanical errors. By continuously measuring and adjusting the pipette’s position in real time, the closed-loop controller minimized alignment errors, even in the presence of miscalibrations or parameter offsets. This approach ensured robust and consistent performance, allowing the robot to precisely insert the pipette into the wells of an arbitrary placed 96-well plate. 

This precision ensured consistent positioning throughout a 15-hour yeast growth experiment. While splitting was manually initiated as a safety precaution, RoboCulture accurately tracked the growth of the three replicates, determining the optimal splitting time based on the declining slope of the growth curve extracted from the monitoring system. The newly split wells exhibited growth consistent with the initial seeding density of their parent wells, confirming successful splitting. However, relying on visual servoing for guidance requires highly accurate perception to prevent unintended robot motion due to perception errors.

Our pipette tip exchange mechanism has been essential for continuous operation, enabling reliable, hands-free tip replacement while reducing contamination risks. Likewise, the integrated growth monitoring system provided real-time feedback on cell culture progress, allowing the system to make informed decisions and further enhancing its overall autonomy and robustness. The experimental setup allows for flexible arrangements, since positions of fixed objects are represented with AprilTags and corresponding offsets. 

We implemented a behavior tree to enable reactive decision-making, allowing the system to adapt dynamically during execution. For instance, a perception failure could result in unsafe robot motion, compromising the experiment. Instead of following a rigid sequence of actions, the behavior tree enabled the robot to respond appropriately by disabling motion and re-establishing perception when needed. This framework provided a structured way to incorporate real-time feedback, ensuring that the robot could autonomously recover from errors and continue to operate without human intervention.

RoboCulture’s key highlight is flexibility in place of throughput. Our robotic system cannot match the sheer speed delivered by commercial liquid handling devices or even manipulator systems with pre-programmed waypoints, but rather it is designed to accommodate a wide range of tasks and environments. We deliberately choose to sacrifice speed in favor of prioritizing autonomy. We are striving toward a direction where efficiency is recuperated by the system being able to operate uninterrupted, without human intervention for long periods of time. The yeast growth experiment is implemented using a collection of parametric behaviors that represent the core actions useful for cell culture. These behaviors can be reused to define custom assays, and the hierarchical structure of the behavior tree enables a modular and scalable representation of experimental workflows.

\subsection*{Limitations and Future Work}

RoboCulture represents a major step toward our vision of generalizable autonomy in biological laboratories, yet its current perception system highlights key limitations that must be addressed to achieve true versatility. Although our visual servoing system delivers highly repeatable pipetting, it is specifically tailored for well plates and does not generalize to other laboratory equipment that demands high-precision positioning. Similarly, our approach for localizing static labware using AprilTags performs well in controlled conditions but assumes that the tags remain unoccluded as the robot moves throughout the workspace. In addition, the tags require a calibration process to determine the relative offset from the tag to the device, which is fragile and time-consuming to maintain. These challenges highlight the need for a more robust and adaptable perception system that can reliably handle a variety of laboratory equipment and experimental conditions without extensive manual calibration. A more general perception system capable of planning complex motions and interacting with diverse labware would represent a substantial improvement over our current approach. It is also important to note that our visual servoing is limited to two dimensions, a constraint that suffices for planar pipetting but falls short for more complex applications, like organoid culture, which require six-degree-of-freedom control. 

Currently, RoboCulture focuses on a specific liquid handling-based assay to demonstrate its capabilities. While the behavior tree framework is modular and can be customized to handle other liquid handling procedures, there is a present limitation on supported tasks, such as well plate manipulation. Although liquid handling is generally considered to be the most ubiquitous aspect of laboratory automation, alternative approaches emphasize flexibility and robustness by addressing tasks beyond pipetting. Some robot manipulator setups deliberately focus on the steps outside of liquid handling such as manipulation of labware, transferring samples between devices and retrieving growth media from refrigerator, since these are often neglected yet crucial for complete end-to-end lab automation \cite{Zwirnmann_2023, pai2024precise, makarova2024lucidgrasproboticframeworkautonomous}. Other perception enhancements such as integrating success/failure detection into the behavior tree and incorporating vision-language model (VLM) capabilities to enable more open-ended task solving are promising directions.

Although RoboCulture is not yet a fully generalizable system, the proposed improvements mark a promising step toward a truly versatile platform for end-to-end laboratory automation. While the expansion of suspended yeast cell culture may be considered a relatively simple task with limited biological significance, demonstrating full autonomy in a standard cell culture lab setting highlights RoboCulture’s potential to alleviate overtime work, reduce the burden of repetitive procedures, and enhance consistency and reproducibility. Its modularity and adaptability allow it to perform a range of tasks across different days, accommodating the dynamic nature of experimental workflows. RoboCulture is not intended to function as a high-throughput public transit system designed to serve thousands simultaneously. Rather, it operates more like an autonomous vehicle tailored to individual or small-group use. Specifically, it is engineered to alleviate repetitive and time-consuming tasks within the laboratory environment, thereby enhancing the daily workflow and improving quality of life for research personnel. 
\section{Experimental Procedures}

\subsection{Resource Availability}
\subsubsection{Lead contact}
Further information and requests for resources and reagents should be directed to and will be fulfilled by the lead contact, Florian Shkurti \mbox{(florian@cs.toronto.edu)}.

\subsubsection{Materials Availability}
This study did not generate new unique materials.

\subsubsection{Data and Code Availability}
The code for RoboCulture and data generated from the experiment can be found on \href{https://ac-rad.github.io/roboculture/}{https://ac-rad.github.io/roboculture/}.

\subsection{Digital Pipette v2}

\begin{figure}
\centering
\includegraphics[width=1.0\linewidth]{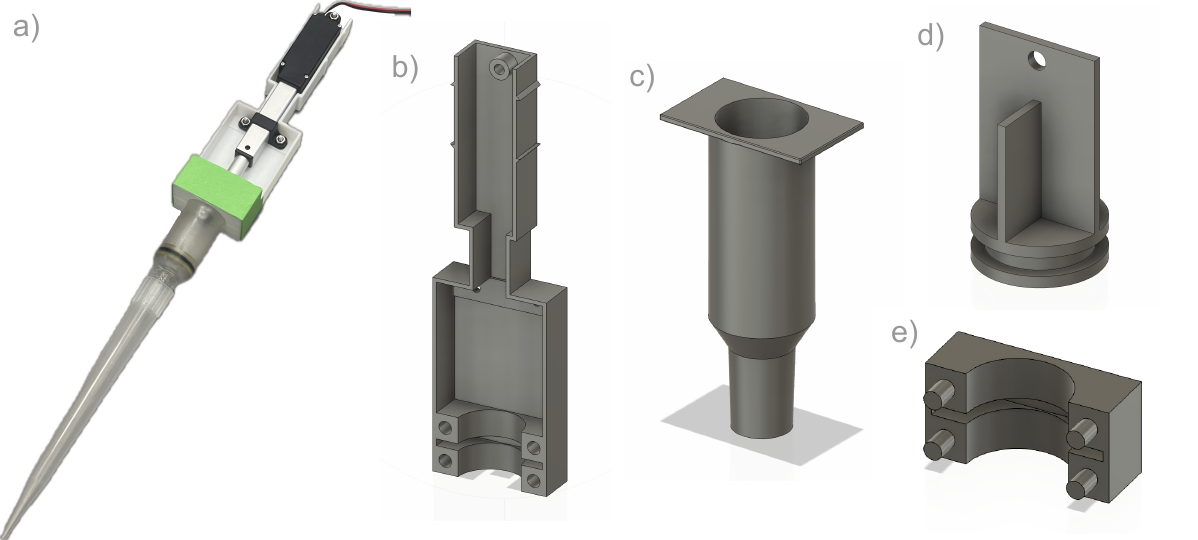}
\caption{(a) Photo of the assembled Digital Pipette v2. (b–e) 3D CAD models of the Digital Pipette v2 components: (b) platform, (c) syringe, (d) plunger, and (e) cover}
\label{fig:pipette}
\end{figure}

The Digital Pipette v2 consists of four 3D-printed components: the platform, syringe, cover, and plunger. A 5cm linear actuator (L16-50-63-6-R, Actuonix), secured to the platform using screws and a mounting bracket, pulls on the plunger to generate suction inside the pipette tip. An O-ring ensures an airtight seal within the syringe piece, while grease is used to maintain an airtight connection between the pipette tip and the syringe body. The cover holds all components securely in place. An Arduino microcontroller interfaces with the robot workstation via USB-serial communication to control the linear actuator, allowing for interoperability with a robot gripper without extensive hardware modifications. A photo of the assembled pipette alongside the CAD models is shown in \textbf{Figure \ref{fig:pipette}}.

The Digital Pipette v2 can be assembled in under 10 minutes, with detailed assembly instructions and CAD models available in our GitHub repository. The 3D models of the Digital Pipette v2 were designed with Fusion 360 (Autodesk Inc.). The platform and the cover were printed with white ANYCUBIC PLA using a KP3S printer (KINGROON Tech Co., Ltd), and the syringe and plunger pieces were printed with Clear Resin V4 (Formlabs Inc.) using a Form 3L printer (Formlabs Inc.).

We use an Uno 328 AVR Dev Board (Creatron Inc., Canada), which is compatible with the Arduino Uno Rev3, to control a linear actuator (L16-50-63-6-R, Actuonix, Canada). The actuator is powered by a 6V DC supply and communicates with the controller PC through USB serial. The linear actuator in the Digital Pipette v2 is operated by a 5V signal, where the pulse length determines the extension of the actuator. In order to derive a relationship between pulse length (in microseconds) and delivered volume (in milliliters), the pipette was calibrated around target volumes. The calibration plots are shown in \textbf{Figures \ref{fig:calib-all}-\ref{fig:calib-01}}.

The approximate prices for the components used to build the Digital Pipette v2 are shown in \textbf{Table \ref{tab:pipette-cost}}. The 3D printing prices were estimated using online printing services (Xometry, Shapeways). Our pipette can be built for under 300 USD, which falls within the typical price range of laboratory micropipettes, which generally cost between 50 and 350 USD.

\begin{table}[h]
    \centering
    \begin{tabular}{l l l}
        \toprule
        \textbf{Parts} & \textbf{Price (USD)} \\
        \midrule
        Linear actuator (Actuonix L16-50-63-6-R) & 70 \\
        Est. 3D printing cost for PLA parts (Platform, Cover) & 35 \\
        Est. 3D printing cost for resin parts (Syringe, Plunger) & 125 \\
        Electronic parts (Arduino, cables, connectors) & 40 \\
        \bottomrule
    \end{tabular}
    \caption{Cost analysis of our proposed Digital Pipette v2.}
\label{tab:pipette-cost}
\end{table}

\subsection{Visual Servoing for Liquid Handling}

\subsubsection{Pipette Tip Detection}
The grasped pipette and camera are both attached to the robot’s gripper, keeping the pipette fixed within the camera frame. However, slight variations in the grasping and tip attachment processes introduce uncertainty in the pipette tip's location within the image, meaning it is infeasible to hard-code the tip coordinates. To address this, we assume that points near the top of the image, where the pipette is wider and closer to the fingers of the gripper, reliably belong to the pipette itself. We select two such points to be used as ``point prompts" for FastSAM \cite{fastsam}, a segmentation model that produces a binary mask of the object containing the prompt points. Using the mask of the pipette, the coordinates of the tip can be determined by identifying the point(s) on the mask with the lowest y-coordinate. If there are multiple points, the x-coordinates of these points are averaged. The pipette tip detection process is shown in \textbf{Figure \ref{fig:perception}a}. Since the pipette remains fixed within the camera frame, tip detection is only required when a new tip is attached.

\subsubsection{Well Detection}
To align the pipette with any well of a 96-well plate, all 96 wells must be detected in real time with high precision. Since well plate dimensions are standardized, accurately detecting the edges of the plate allows us to infer the positions of the individual wells relative to the plate. 
For the initial frame, we use FastSAM \cite{fastsam} with the text prompt ``microplate" to produce a binary mask of the well plate (\textbf{Figure \ref{fig:perception}a}). While FastSAM reports near-realtime inference speeds with optimal hardware, we instead rely on lightweight tracking methods to keep the system hardware-agnostic. 

First, we preprocess the raw image by applying Gaussian blur for high-frequency noise reduction, followed by normalization and contrast enhancement. We then apply Canny edge detection \cite{canny} to obtain a binary representation of the image. We then use the mask of the well plate obtained from FastSAM to mask out edges that do not belong to the well plate. Additionally, the mask of the pipette is used to remove edges belonging to the pipette. Next, we determine the well plate's bounding rectangle using OpenCV \cite{opencv_library}, extracting its four corner points (\textbf{Figure \ref{fig:perception}b}). These detected corners allow us to compute a projective transformation relative to a predefined ``template" of well plate dimensions in millimeters. This projective transform can then be applied to the template well points to obtain their perceived locations in the image. 
The well detection process is shown in \textbf{Figure \ref{fig:perception}b, c, d}. To track the well plate across frames, we use the bounding rectangle from the previous frame as a mask and only consider edges within the mask plus some padding. If the proportions of the bounding rectangle deviates significantly from expected plate proportions, the track is considered failed, and the plate is re-segmented to restore the mask. The track is prone to fail in certain situations where there are temporary occlusions of the well plate. Robot motion is disabled whenever the perception pipeline fails and resumes once the plate is re-located. 

\subsubsection{Feedback Controller}
Once the coordinates of the wells and the pipette tip are identified in the image, we define an error vector as the difference between their pixel coordinates:

\begin{equation}
e = (x_{\text{tip}} - x_{\text{well}}, y_{\text{tip}} - y_{\text{well}})
\end{equation}

The objective is to minimize $e$ by adjusting the robot end-effector's pose incrementally. This is accomplished by computing a small incremental movement in the direction of $e$ and continuously updating the robot’s pose until alignment is reached. Algorithm \ref{alg:servo_control} outlines this process, which takes as input the latest error vector $e\_vec$ obtained from the perception pipeline, the current Cartesian robot end-effector pose $curr\_pose$, a proportional control gain $k_p$, a control signal limit $u\_lim$, and an image error threshold $img\_thresh$. 

First, a control signal is computed as the image error vector scaled by the gain. The x and y components of this signal are then swapped to align the image axes with the robot’s base frame. While this approach does not explicitly incorporate the full camera-to-robot transformation, the closed-loop nature of the system ensures that incremental corrections continuously refine the alignment, compensating for errors over successive iterations. To prevent excessive movements, the control signal is capped at  $u\_lim$  when its magnitude exceeds the limit. The updated pose is then sent to the robot, where the internal controller computes the necessary joint angles via inverse kinematics. To enable smooth, real-time adjustments, we utilize FrankaPy’s \cite{zhang2020modular} dynamic control mode, which allows for streaming relative Cartesian poses instead of executing a single blocking trajectory.

By dynamically adjusting the pose rather than estimating a single Cartesian destination based on image coordinates, the system can adapt to minor shifts in plate position and perception errors. Additionally, due to imperfections in camera extrinsic calibration, a single pose adjustment is often insufficient, making iterative refinement necessary. The visual servoing control operates exclusively in the XY plane of the end-effector, assuming a fixed plate height. As a result, the Z component of the end-effector can be controlled separately. By restricting visual servoing to two dimensions without altering camera orientation, we eliminate the need for an explicit image Jacobian, significantly simplifying the control strategy.

\begin{algorithm}
\caption{Servo Control Algorithm}\label{alg:servo_control}
\begin{algorithmic}[1]
\Require ${e\_vec}$, ${curr\_pose}$, $k_p$, ${u\_lim}$, ${img\_thresh}$
\While{True}
    \State $u = k_p \cdot {e\_vec}$ 
    \State $u \gets [u_1, u_0, 0]$ 
    \If{$\lvert u \rvert > {u\_lim} $}
        \State $u \gets u \cdot ({u\_lim} / \lvert u \rvert)$ 
    \EndIf

    \State $curr\_pose \gets curr\_pose + u$

    \State Send ${curr\_pose}$ to robot

    \If{$\lvert {e\_vec} \rvert < {img\_thresh}$}
        \State \textbf{Break}.
    \EndIf
    \State Sleep for the control rate
    
\EndWhile
\end{algorithmic}
\end{algorithm}

\subsection{Optical Density Perception for Growth Monitoring}
We measured brightness by converting each image to the HSV color space and extracting the value (V) channel. The brightness values from each of the six replicates for a particular group are then averaged and passed through a rolling average smoothing operation, producing a single data point per group for each time step. By analyzing the smoothed derivative of these data points, we estimate when the growth curve begins to plateau, signaling that a group has reached saturation. The growth monitoring procedure is visualized in \textbf{Figure \ref{fig:process}b}. Because the RGB camera must capture a direct view of the well bottoms, it cannot image all wells simultaneously without also capturing the sides of the wells, which would interfere with accurate measurements. To ensure each well is imaged correctly, the robot moved the camera across the plate, capturing images in eight separate patches, each containing a 4×3 grid of wells, resulting in 96 individual images, each 24×24 pixels in size. 

\subsection{Behavior Tree and Experiment State Handling}
The \verb|py_trees_ros| package was used in conjunction with ROS Noetic to implement the behavior tree framework. 
Descriptions of the behaviors developed for RoboCulture as well as components of the behavior tree used for the yeast growth experiment are shown in Section \ref{sec:behavior-trees}. 
To further enhance system adaptability, we implemented a mechanism for real-time parameter tuning through an rqt-based graphical user interface (\textbf{Figure \ref{fig:rqt}}). This interface allows both the user and the behavior tree to adjust parameters such as pipette actuator length or the specific well numbers to access during execution. When a parameter is updated, the behavior tree automatically registers the change and applies it during the next execution cycle, ensuring seamless integration into the ongoing workflow. This setup offers an intuitive means for modifying experimental procedures on the fly, supporting rapid protocol development and testing. It also streamlines debugging by enabling developers to explore different parameter configurations in real time without altering the underlying code. An image of the rqt interface and a description of the tunable parameters is provided in Section \ref{sec:rqt}.

\subsection{Yeast Experiment}

\subsubsection{Setup}

\textbf{Figure \ref{fig:setup}} shows the experimental setup for the yeast growth experiment, including our Franka Emika robot with a Robotiq 2F-85 gripper and attached Intel RealSense D435i camera, an OHAUS SHHD1619DG Heavy Duty Orbital Shaker Platform with the well plate on top. In addition, the scene consists of various supporting laboratory materials including the Digital Pipette v2, and stationary objects including a 3D printed pipette tip remover, a 3D printed
pipette tip rack, a biological waste bin and a Falcon tube rack. AprilTags~\cite{olson2011tags} were used to denote the positions of the stationary objects. CAD models for the pipette tip rack and pipette tip remover are shown in \textbf{Figures \ref{fig:tiprack}-\ref{fig:tipremover}}.

Our Digital Pipette v2, 2) a downwards-facing Intel RealSense D435i camera,
3) a Franka Emika robot with a Robotiq 2F-85 gripper, 4) an OHAUS SHHD1619DG Heavy Duty Orbital Shaker Platform,
5) a 3D printed pipette tip remover, 6) a biological waste bin, 7) a Falcon Tube rack holding YPD media, 8) a 3D printed
pipette tip rack, and 9) a 96 well plate prepared with yeast.

\subsubsection{Sample Preparation}

Saccharomyces cerevisiae was maintained on agarose plates containing 2\% agarose in YPD broth (Gibco, cat \#A1374501) at 4ºC under sterile conditions. Cell expansion was carried out as described in \cite{Olivares-Marin2018}. In short, 5 mL of YPD broth (Gibco, cat \#A1374501) was inoculated in a sterile round-bottom tube and incubated overnight at 30ºC to promote cell growth. The cells were then collected and counted using a hemocytometer. Serial dilutions in YPD broth were prepared to achieve final concentrations of 50 million cells/mL, 30 million cells/mL, and 10 million cells/mL. The prepared cell suspensions were dispensed into two identical 96-well plates (Grenier) using a multichannel pipette. One plate was handled by the robot, while the other served as a reference for optical density measurements over time. The reference plate was placed in a plate reader, where optical density was measured every five minutes at 600 nm, with orbital shaking preceding each measurement.

\section*{Acknowledgments}
We would like to thank Yimu Zhao for her insightful discussions. This research was undertaken thanks in part to funding provided to the University of Toronto's Acceleration Consortium from the Canada First Research Excellence Fund, grant number CFREF-2022-00042. S.O. is supported by NSERC CGS and Additional Ventures Grant.

\section*{Author contributions}
K.A. led, developed, and validated the system.
K.D. co-led the technical development and validation of the system.
N.Y. co-developed Digital Pipette v2.
S.O. provided the cell culturing procedure, prepared the initial samples, and performed the plate reader growth monitoring experiment.
D.B. contributed to the cell culturing procedure. 
S.O. and D.B. performed the human pipetting experiment.
I.Y. contributed to the optical density perception algorithm and discussions on existing liquid handling systems.
A.A.G., F.S., and M.R. supervised the project.
K.A., K.D., S.O., F.S., and M.R. wrote the initial draft and revised the manuscript.
All authors proofread the manuscript.

\section*{Declaration of interests}
The authors declare no competing interests.

\bibliography{Bibliography/Bibliography}

\begin{thebibliography}{10}
\providecommand{\url}[1]{#1}
\csname url@samestyle\endcsname
\providecommand{\newblock}{\relax}
\providecommand{\bibinfo}[2]{#2}
\providecommand{\BIBentrySTDinterwordspacing}{\spaceskip=0pt\relax}
\providecommand{\BIBentryALTinterwordstretchfactor}{4}
\providecommand{\BIBentryALTinterwordspacing}{\spaceskip=\fontdimen2\font plus
\BIBentryALTinterwordstretchfactor\fontdimen3\font minus \fontdimen4\font\relax}
\providecommand{\BIBforeignlanguage}[2]{{%
\expandafter\ifx\csname l@#1\endcsname\relax
\typeout{** WARNING: IEEEtran.bst: No hyphenation pattern has been}%
\typeout{** loaded for the language `#1'. Using the pattern for}%
\typeout{** the default language instead.}%
\else
\language=\csname l@#1\endcsname
\fi
#2}}
\providecommand{\BIBdecl}{\relax}
\BIBdecl

\bibitem{tom2024self}
\BIBentryALTinterwordspacing
G.~Tom, S.~P. Schmid, S.~G. Baird, Y.~Cao, K.~Darvish, H.~Hao, S.~Lo, S.~Pablo-García, E.~M. Rajaonson, M.~Skreta, N.~Yoshikawa, S.~Corapi, G.~D. Akkoc, F.~Strieth-Kalthoff, M.~Seifrid, and A.~Aspuru-Guzik, ``Self-driving laboratories for chemistry and materials science,'' \emph{Chemical Reviews}, vol. 124, no.~16, pp. 9633--9732, aug 2024. [Online]. Available: \url{https://doi.org/10.1021/acs.chemrev.4c00055}
\BIBentrySTDinterwordspacing

\bibitem{hase2019nextgenlabs}
\BIBentryALTinterwordspacing
F.~Häse, L.~M. Roch, and A.~Aspuru-Guzik, ``Next-generation experimentation with self-driving laboratories,'' \emph{Trends in Chemistry}, vol.~1, no.~3, pp. 282--291, 2019. [Online]. Available: \url{https://doi.org/10.1016/j.trechm.2019.02.007}
\BIBentrySTDinterwordspacing

\bibitem{opentrons_ot2}
\BIBentryALTinterwordspacing
O.~L. Inc., ``Ot-2 liquid handling robot,'' 2024, accessed: 2024-11-11. [Online]. Available: \url{https://opentrons.com/robots/ot-2}
\BIBentrySTDinterwordspacing

\bibitem{hamilton_star_v}
\BIBentryALTinterwordspacing
H.~Company, ``Microlab star v automated liquid handling platform,'' 2024, accessed: 2024-11-11. [Online]. Available: \url{https://www.hamiltoncompany.com/automated-liquid-handling/platforms/microlab-star-v}
\BIBentrySTDinterwordspacing

\bibitem{flowbot_one}
\BIBentryALTinterwordspacing
F.~Robotics, ``Flowbot® one: The intuitive liquid handling robot,'' 2024, accessed: 2024-11-11. [Online]. Available: \url{https://flow-robotics.com/products/flowbot-one/}
\BIBentrySTDinterwordspacing

\bibitem{elanzew2020stemcellfactory}
\BIBentryALTinterwordspacing
A.~Elanzew \emph{et~al.}, ``The stemcellfactory: A modular system integration for automated generation and expansion of human induced pluripotent stem cells,'' \emph{Frontiers in Bioengineering and Biotechnology}, vol.~8, p. 580352, 2020. [Online]. Available: \url{https://doi.org/10.3389/fbioe.2020.580352}
\BIBentrySTDinterwordspacing

\bibitem{hamilton_cell_care_star}
\BIBentryALTinterwordspacing
H.~Company, ``Cell care star: Automated cell culture system,'' 2024, accessed: 2024-11-11. [Online]. Available: \url{https://www.hamiltoncompany.com/automated-liquid-handling/assay-ready-workstations/cell-care-star}
\BIBentrySTDinterwordspacing

\bibitem{holland2020lifescienceautomation}
\BIBentryALTinterwordspacing
I.~Holland and J.~A. Davies, ``Automation in the life science research laboratory,'' \emph{Frontiers in Bioengineering and Biotechnology}, vol.~8, 2020. [Online]. Available: \url{https://doi.org/10.3389/fbioe.2020.571777}
\BIBentrySTDinterwordspacing

\bibitem{cooper2025accelerating}
\BIBentryALTinterwordspacing
A.~I. Cooper, P.~Courtney, K.~Darvish, M.~Eckhoff, H.~Fakhruldeen, A.~Gabrielli, A.~Garg, S.~Haddadin, K.~Harada, J.~Hein, M.~Hübner, D.~Knobbe, G.~Pizzuto, F.~Shkurti, R.~Shrestha, K.~Thurow, R.~Vescovi, B.~Vogel-Heuser, Ádám Wolf, N.~Yoshikawa, Y.~Zeng, Z.~Zhou, and H.~Zwirnmann, ``Accelerating discovery in natural science laboratories with ai and robotics: Perspectives and challenges from the 2024 ieee icra workshop, yokohama, japan,'' 2025. [Online]. Available: \url{https://doi.org/10.48550/arXiv.2501.06847}
\BIBentrySTDinterwordspacing

\bibitem{angelopoulos2024transforming}
\BIBentryALTinterwordspacing
A.~Angelopoulos, J.~F. Cahoon, and R.~Alterovitz, ``Transforming science labs into automated factories of discovery,'' \emph{Science Robotics}, vol.~9, no.~95, p. eadm6991, 2024. [Online]. Available: \url{https://doi.org/10.1126/scirobotics.adm6991}
\BIBentrySTDinterwordspacing

\bibitem{zhang2022integratingmanualpipettecollaborative}
\BIBentryALTinterwordspacing
J.~Zhang, W.~Wan, N.~Tanaka, M.~Fujita, and K.~Harada, ``Integrating a manual pipette into a collaborative robot manipulator for flexible liquid dispensing,'' 2022. [Online]. Available: \url{https://doi.org/10.1109/TASE.2023.3312657}
\BIBentrySTDinterwordspacing

\bibitem{cellcultureautomation_ais}
\BIBentryALTinterwordspacing
J.~Hamm, S.~Lim, J.~Park, J.~Kang, I.~Lee, Y.~Lee, J.~Kang, Y.~Jo, J.~Lee, S.~Lee, M.~C. Ratri, A.~I. Brilian, S.~Lee, S.~Jeong, and K.~Shin, ``A modular robotic platform for biological research: Cell culture automation and remote experimentation,'' \emph{Advanced Intelligent Systems}, vol.~6, no.~5, p. 2300566, 2024. [Online]. Available: \url{https://doi.org/10.1002/aisy.202300566}
\BIBentrySTDinterwordspacing

\bibitem{firoozi2023foundation}
R.~Firoozi, J.~Tucker, S.~Tian, A.~Majumdar, J.~Sun, W.~Liu, Y.~Zhu, S.~Song, A.~Kapoor, K.~Hausman \emph{et~al.}, ``Foundation models in robotics: Applications,'' \emph{Challenges, and the Future}, 2023.

\bibitem{digital-pipette}
\BIBentryALTinterwordspacing
N.~Yoshikawa, K.~Darvish, M.~G. Vakili, A.~Garg, and A.~Aspuru-Guzik, ``Digital pipette: open hardware for liquid transfer in self-driving laboratories,'' \emph{Digital Discovery}, vol.~2, pp. 1745--1751, 2023. [Online]. Available: \url{http://doi.org/10.1039/D3DD00115F}
\BIBentrySTDinterwordspacing

\bibitem{colledanchisebtrees}
M.~Colledanchise and P.~Ogren, \emph{Behavior Trees in Robotics and AI: An Introduction}, 07 2018.

\bibitem{olson2011tags}
E.~Olson, ``{AprilTag}: A robust and flexible visual fiducial system,'' in \emph{Proceedings of the {IEEE} International Conference on Robotics and Automation ({ICRA})}.\hskip 1em plus 0.5em minus 0.4em\relax IEEE, May 2011, pp. 3400--3407.

\bibitem{ISO8655-2_2022}
{International Organization for Standardization}, ``Piston-operated volumetric apparatus – part 2: Pipettes,'' International Organization for Standardization, Tech. Rep. ISO 8655-2:2022(en), 2022.

\bibitem{ISO8655-6_2022}
------, ``Piston-operated volumetric apparatus – part 6: Gravimetric reference measurement procedure for the determination of volume,'' International Organization for Standardization, Tech. Rep. ISO 8655-6:2022(en), 2022.

\bibitem{sri1979real}
\BIBentryALTinterwordspacing
S.~I. A.~I. Center and G.~Agin, \emph{Real time control of a robot with a mobile camera}, ser. Technical note.\hskip 1em plus 0.5em minus 0.4em\relax SRI International, 1979. [Online]. Available: \url{https://books.google.ca/books?id=XEAtGwAACAAJ}
\BIBentrySTDinterwordspacing

\bibitem{Beal2020}
\BIBentryALTinterwordspacing
J.~Beal \emph{et~al.}, ``Robust estimation of bacterial cell count from optical density,'' \emph{Communications Biology}, vol.~3, no.~1, p. 512, 2020. [Online]. Available: \url{https://doi.org/10.1038/s42003-020-01127-5}
\BIBentrySTDinterwordspacing

\bibitem{mira2022estimating}
\BIBentryALTinterwordspacing
P.~Mira, P.~Yeh, and B.~G. Hall, ``Estimating microbial population data from optical density,'' \emph{PLOS ONE}, vol.~17, no.~10, pp. 1--8, 10 2022. [Online]. Available: \url{https://doi.org/10.1371/journal.pone.0276040}
\BIBentrySTDinterwordspacing

\bibitem{Zwirnmann_2023}
\BIBentryALTinterwordspacing
H.~Zwirnmann, D.~Knobbe, U.~Culha, and S.~Haddadin, ``Towards flexible biolaboratory automation: Container taxonomy-based, 3d-printed gripper fingers*,'' in \emph{2023 IEEE/RSJ International Conference on Intelligent Robots and Systems (IROS)}.\hskip 1em plus 0.5em minus 0.4em\relax IEEE, Oct. 2023, p. 6823–6830. [Online]. Available: \url{http://doi.org/10.1109/IROS55552.2023.10342218}
\BIBentrySTDinterwordspacing

\bibitem{pai2024precise}
S.~Pai, K.~Takahashi, S.~Masuda, N.~Fukaya, K.~Yamane, and A.~Ummadisingu, ``Precise well-plate placing utilizing contact during sliding with tactile-based pose estimation for laboratory automation,'' in \emph{2024 IEEE/RSJ International Conference on Intelligent Robots and Systems (IROS)}.\hskip 1em plus 0.5em minus 0.4em\relax IEEE, 2024.

\bibitem{makarova2024lucidgrasproboticframeworkautonomous}
\BIBentryALTinterwordspacing
M.~Makarova, D.~Trinitatova, Q.~Liu, and D.~Tsetserukou, ``Lucidgrasp: Robotic framework for autonomous manipulation of laboratory equipment with different degrees of transparency via 6d pose estimation,'' 2024. [Online]. Available: \url{https://doi.org/10.48550/arXiv.2410.07801}
\BIBentrySTDinterwordspacing

\bibitem{fastsam}
\BIBentryALTinterwordspacing
X.~Zhao, W.~Ding, Y.~An, Y.~Du, T.~Yu, M.~Li, M.~Tang, and J.~Wang, ``Fast segment anything,'' 2023. [Online]. Available: \url{https://doi.org/10.48550/arXiv.2306.12156}
\BIBentrySTDinterwordspacing

\bibitem{canny}
J.~Canny, ``A computational approach to edge detection,'' \emph{IEEE Transactions on Pattern Analysis and Machine Intelligence}, vol. PAMI-8, no.~6, pp. 679--698, 1986.

\bibitem{opencv_library}
G.~Bradski, ``{The OpenCV Library},'' \emph{Dr. Dobb's Journal of Software Tools}, 2000.

\bibitem{zhang2020modular}
K.~Zhang, M.~Sharma, J.~Liang, and O.~Kroemer, ``A modular robotic arm control stack for research: Franka-interface and frankapy,'' \emph{arXiv preprint arXiv:2011.02398}, 2020.

\bibitem{Olivares-Marin2018}
I.~K. Olivares-Marin, J.~C. González-Hernández, C.~Regalado-Gonzalez, and L.~A. Madrigal-Perez, ``Saccharomyces cerevisiae exponential growth kinetics in batch culture to analyze respiratory and fermentative metabolism,'' \emph{JoVE}, no. 139, p. e58192, 2018.

\bibitem{garrido-jurado2014automatic}
S.~Garrido-Jurado, R.~Muñoz-Salinas, F.~J. Madrid-Cuevas, and M.~J. Marín-Jiménez, ``Automatic generation and detection of highly reliable fiducial markers under occlusion,'' \emph{Pattern Recognition}, vol.~47, no.~6, pp. 2280--2292, 2014.

\bibitem{knobbe2022core}
D.~Knobbe, H.~Zwirnmann, M.~Eckhoff, and S.~Haddadin, ``Core processes in intelligent robotic lab assistants: Flexible liquid handling,'' in \emph{2022 IEEE/RSJ International Conference on Intelligent Robots and Systems (IROS)}.\hskip 1em plus 0.5em minus 0.4em\relax IEEE, 2022.

\bibitem{angelopoulos2023high}
A.~Angelopoulos, M.~Verber, C.~McKinney, J.~Cahoon, and R.~Alterovitz, ``High-accuracy injection using a mobile manipulation robot for chemistry lab automation,'' in \emph{2023 IEEE/RSJ International Conference on Intelligent Robots and Systems (IROS)}, 2023, pp. 10\,102--10\,109.

\bibitem{schober2024vision}
\BIBentryALTinterwordspacing
D.~Schober, R.~Güldenring, J.~Love, and L.~Nalpantidis, ``Vision-based robot manipulation of transparent liquid containers in a laboratory setting,'' \emph{arXiv preprint arXiv:2404.16529}, 2024. [Online]. Available: \url{https://doi.org/10.48550/arXiv.2404.16529}
\BIBentrySTDinterwordspacing

\bibitem{kennedy2017precise}
M.~Kennedy, K.~Queen, D.~Thakur, K.~Daniilidis, and V.~Kumar, ``Precise dispensing of liquids using visual feedback,'' in \emph{2017 IEEE/RSJ International Conference on Intelligent Robots and Systems (IROS)}, 2017, pp. 1260--1266.

\bibitem{brooks_preciseflex_3400}
\BIBentryALTinterwordspacing
B.~Automation, ``Preciseflex 3400 collaborative robot,'' 2025, accessed: 2025-01-27. [Online]. Available: \url{https://www.brooks.com/industrial-automation/collaborative-robots/preciseflex-3400/}
\BIBentrySTDinterwordspacing

\bibitem{thermofisher_spinnaker_microplate_robot}
\BIBentryALTinterwordspacing
T.~F. Scientific, ``Spinnaker microplate robot,'' 2025, accessed: 2025-01-27. [Online]. Available: \url{https://www.thermofisher.com/order/catalog/product/SPK0001}
\BIBentrySTDinterwordspacing

\bibitem{openlh}
\BIBentryALTinterwordspacing
G.~Gome, J.~Waksberg, A.~Grishko, I.~Y. Wald, and O.~Zuckerman, ``Openlh: Open liquid-handling system for creative experimentation with biology,'' in \emph{Proceedings of the Thirteenth International Conference on Tangible, Embedded, and Embodied Interaction}, ser. TEI '19.\hskip 1em plus 0.5em minus 0.4em\relax New York, NY, USA: Association for Computing Machinery, 2019, p. 55–64. [Online]. Available: \url{https://doi.org/10.1145/3294109.3295619}
\BIBentrySTDinterwordspacing

\bibitem{ker2011engineered}
\BIBentryALTinterwordspacing
D.~F.~E. Ker, L.~E. Weiss, S.~N. Junkers, M.~Chen, Z.~Yin, M.~F. Sandbothe, S.-i. Huh, S.~Eom, R.~Bise, E.~Osuna-Highley, T.~Kanade, and P.~G. Campbell, ``An engineered approach to stem cell culture: Automating the decision process for real-time adaptive subculture of stem cells,'' \emph{PLOS ONE}, vol.~6, no.~11, pp. 1--12, 11 2011. [Online]. Available: \url{https://doi.org/10.1371/journal.pone.0027672}
\BIBentrySTDinterwordspacing

\end{thebibliography}
\bibliographystyle{IEEEtran}

\newpage
\onecolumn
\section{Supplemental information}

\renewcommand{\thefigure}{S\arabic{figure}}
\setcounter{figure}{0}

\subsection{Related Work}
\label{sec:related-work}

With the common goal of developing more general solutions to laboratory automation, robotic manipulators for biology have been explored. Early approaches often relied on fixed waypoints and manual calibration, limiting their adaptability to variations in experimental setups. More recent efforts integrate computer vision and sensor feedback for real-time adaptation, yet challenges remain in flexibility and robustness across diverse workflows.
\\
\paragraph{Robotic Manipulators for Liquid Handling} 

Recent work has demonstrated that robotic manipulators can achieve high pipette positioning accuracy for randomly placed well plates using collaborative robots~\cite{zhang2022integratingmanualpipettecollaborative}. This approach involves manually calibrating the positions of the pipette tip rack and well plate by guiding the robot to various waypoints along the surfaces of these objects to establish fixed reference points for navigation.
However, this strategy requires recalibration whenever labware is repositioned and often demands additional training to support new object variants, limiting its scalability and adaptability.

Modular robotic platforms for cell culture automation have also been described that use fiducial markers to detect laboratory equipment and position the robot end-effector accordingly \cite{cellcultureautomation_ais}. Here, we also leverage AprilTags to identify positions of larger stationary labware. Yet, for smaller structures such as 96-well plates, fiducial markers like AprilTags \cite{olson2011tags} and ArUco markers \cite{garrido-jurado2014automatic} become highly sensitive to camera calibration, which can result in significant positioning errors. Although coordinate refinement procedures have been proposed to minimize these errors \cite{cellcultureautomation_ais}, the system still relies on a manually defined offset between the marker and the target, such as a pipette tip rack. Additionally, moving the robot end-effector by a fixed offset is inherently open-loop, and no visual feedback is incorporated to ensure the positioning is successful. Although robot motions are highly repeatable, changes in the environment due to protocol variability may warrant tedious calibration of offsets. 

Improving position detection in automated systems has also been achieved using pre-calibrated offsets between ArUco markers and associated labware~\cite{knobbe2022core}, or through the combination of a coarse AprilTag-based localization system with a deep learning visual feedback loop for precise needle positioning in chemistry laboratory automation~\cite{angelopoulos2023high}. Robotic liquid pouring has also been proposed as an alternative to traditional pipetting for liquid handling; however, this approach is not suitable for small-volume applications, particularly in cell culture settings where sterility and precision are critical~\cite{schober2024vision, kennedy2017precise}.

SCARA robots, such as the Brooks PreciseFlex 3400~\cite{brooks_preciseflex_3400} and Thermo Scientific™ Spinnaker™ 3~\cite{thermofisher_spinnaker_microplate_robot}, offer an accessible entry point to laboratory automation. However, their design imposes movement constraints, as motion is limited to a planar surface. This makes them less suitable for tasks that require greater dexterity, such as human-like sample handling or protocols involving tilting and three-dimensional manipulation. To overcome these limitations, we instead employed a 7-axis robotic arm for its versatility, combined AprilTags for localizing larger hardware, and used visual servoing to achieve the precision necessary for identifying and interacting with 96-well plates on a shaking platform.
\\
\paragraph{Robotic Manipulators for Pipette Tip Exchange}

Pipette tip exchange presents another challenge in applications requiring sterility, such as cell culture. One strategy relies on fixed navigation waypoints using a collaborative robot~\cite{zhang2022integratingmanualpipettecollaborative}. For pipette tip attachment, the authors correct residual misalignment by using a vision-based classifier that predicts and corrects positioning errors. The vision feedback strategy is effective, however, it requires re-calibration of the waypoints whenever the positions of the labware change.
Other pipette tip exchange strategies rely on hard coded positions for the end effector \cite{openlh}, or hard coded offsets from a fiducial marker \cite{cellcultureautomation_ais, knobbe2022core} to position the pipette for tip attachment. These methods are rigid and rely on the mechanical precision of the robot, but offer little flexibility when protocols and environments change.
\\
\paragraph{Automation of Cell Growth Monitoring} 

To truly replicate human capabilities, robotic cell culture systems must incorporate automated cell growth monitoring. While in-incubator cameras are often proposed for this purpose, they are typically limited to a single fixed task in a fixed location. Although effective strategies for automating confluency monitoring and supporting decision-making in cell culture have been demonstrated \cite{ker2011engineered}, few methods have been integrated with robotic manipulators as part of a broader laboratory automation workflow.

\FloatBarrier
\newpage
\subsection{Behavior Trees}
\label{sec:behavior-trees}
\subsubsection{Description of Behavior Tree Components}
Behavior trees are a hierarchical control architecture commonly used in robotics and game AI to model complex decision-making processes in a modular and reactive manner. A behavior tree is composed of nodes arranged in a tree structure, where each node represents either a control flow construct (e.g., sequence, selector) or an action (e.g., a task like “move robot” or a check like “is well saturated?”). These actions are referred to as ``behaviors" and can be defined parametrically, making them modular and reusable. 

\begin{figure}[h]
    \centering
    \includegraphics[width=\linewidth]{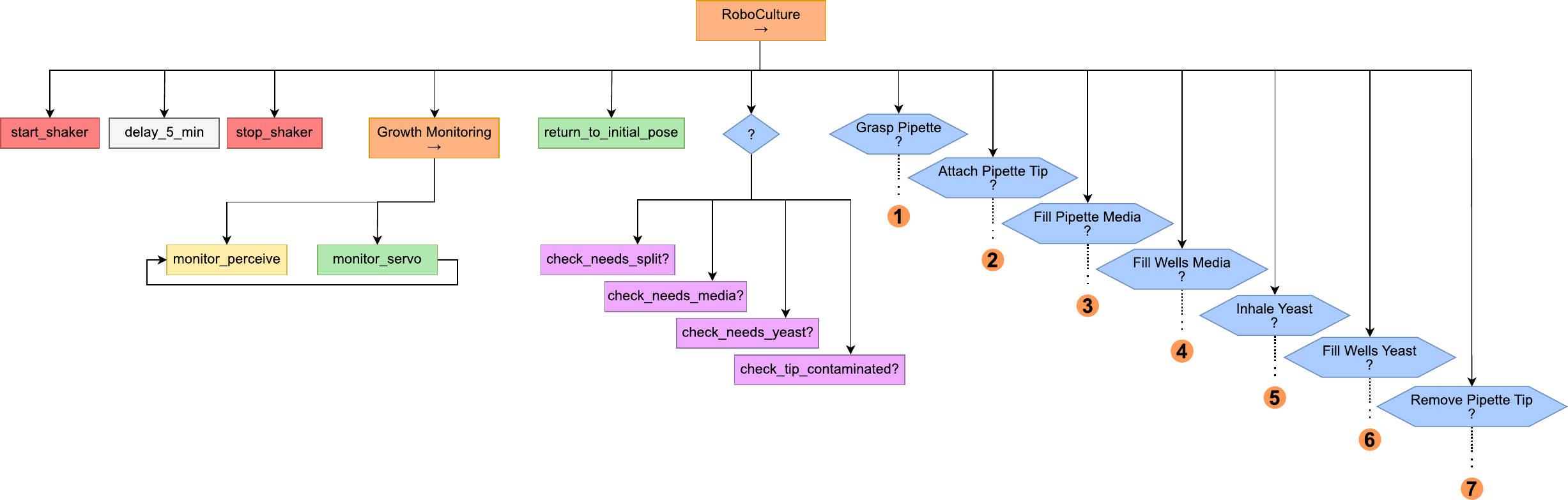}
    \caption{RoboCulture behavior tree. This tree governs the decision-making logic for the 15-hour yeast cell culturing experiment. The leftmost children of the global \texttt{Sequence} manage the periodic enabling and disabling of the shaker every 5 minutes. The growth monitoring \texttt{Sequence} is responsible for imaging the wells in the plate; it includes the \texttt{monitor\_perceive} behavior, which succeeds only on its eighth iteration to simulate a loop and capture images for optical density analysis. The decision to split is determined via parameters in the dynamic reconfigure server: the associated \texttt{Selector} node proceeds only if one of its \texttt{CheckExperimentState} children returns \texttt{SUCCESS}. The remaining seven \texttt{Selector} nodes represent sub-trees that carry out individual experimental tasks. Refer to \textbf{Figures \ref{fig:btree-grasp-pipette}-\ref{fig:btree-remove-tip}} for detailed views of these sub-trees.}
    \label{fig:btree-main}
\end{figure}

\begin{figure}[h]
    \centering
    \includegraphics[width=\linewidth]{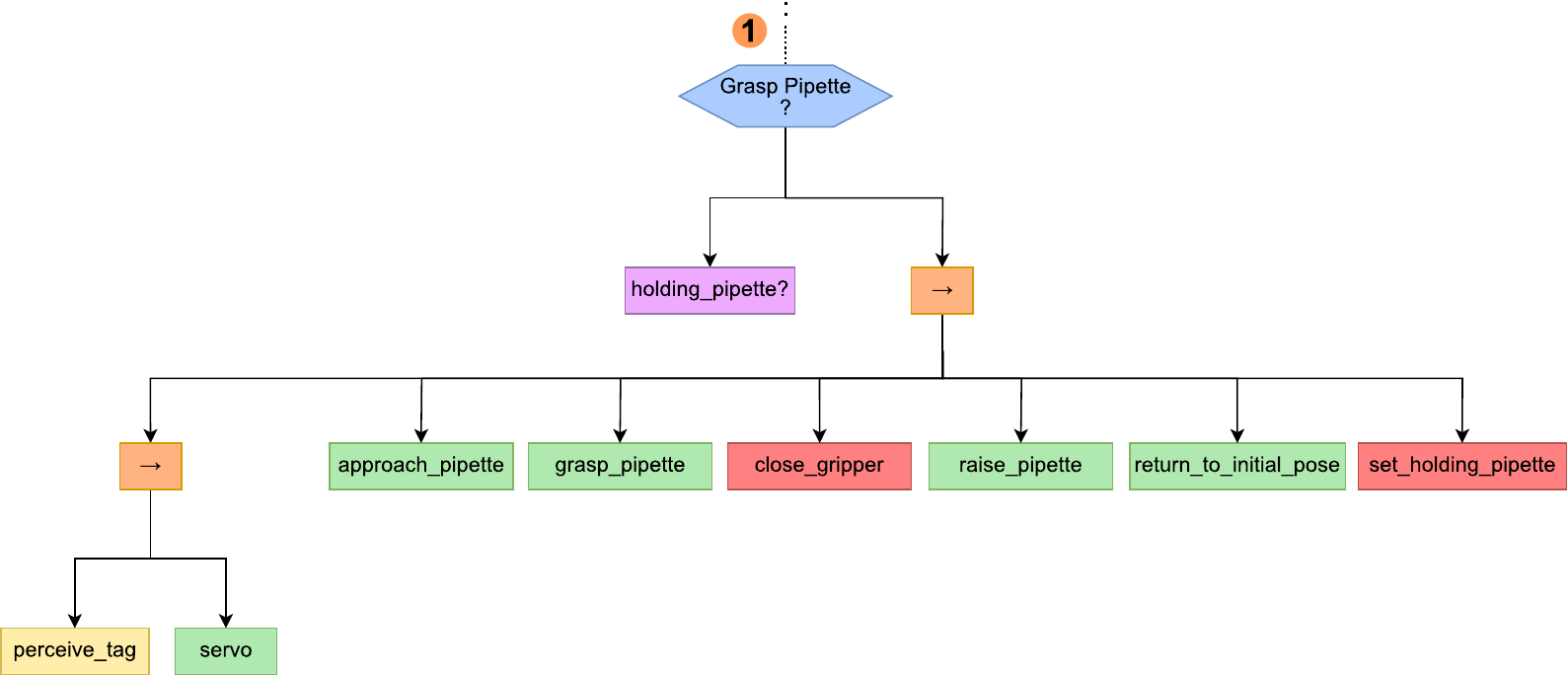}
    \caption{Behavior tree for \texttt{Grasp Pipette}. This sub-tree begins with a \texttt{Selector} node that first checks whether the pipette is already held by evaluating the \texttt{holding\_pipette} condition. If this check fails, the tree proceeds to the right child, a \texttt{Sequence} that performs the full grasping routine. It starts by perceiving the pipette stand’s AprilTag and aligning the end-effector using the \texttt{servo} behavior. The subsequent actions position the robot using predefined positions relative to the AprilTag, as well as control the robot's gripper. The final step sets the internal state to reflect that the pipette is now being held.}
    \label{fig:btree-grasp-pipette}
\end{figure}

\begin{figure}[h]
    \centering
    \includegraphics[width=\linewidth]{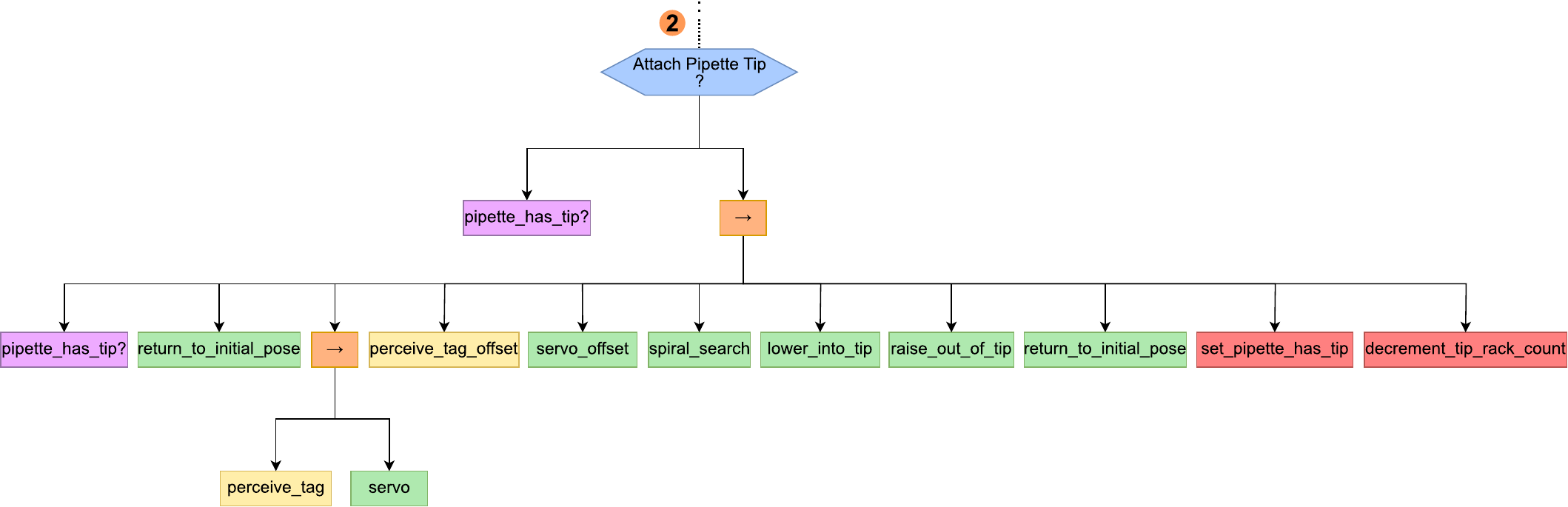}
    \caption{Behavior tree for \texttt{Attach Pipette Tip}. This sub-tree begins with a \texttt{Selector} node that checks whether a tip is already attached by evaluating the \texttt{pipette\_has\_tip} condition. If a tip is already present, the behavior returns \texttt{SUCCESS} immediately. Otherwise, it proceeds to a \texttt{Sequence} that carries out the attachment routine. This includes perceiving the AprilTag of the tip rack with a corresponding offset, and servoing to the target position. A spiral search is used to refine the alignment, followed by lowering the pipette into the tip, lifting it out, and returning to the initial pose. The final steps update the internal experiment state to reflect the new tip status and decrement the remaining tip count in the rack.}
    \label{fig:btree-attach-tip}
\end{figure}

\begin{figure}[h]
    \centering
    \includegraphics[width=\linewidth]{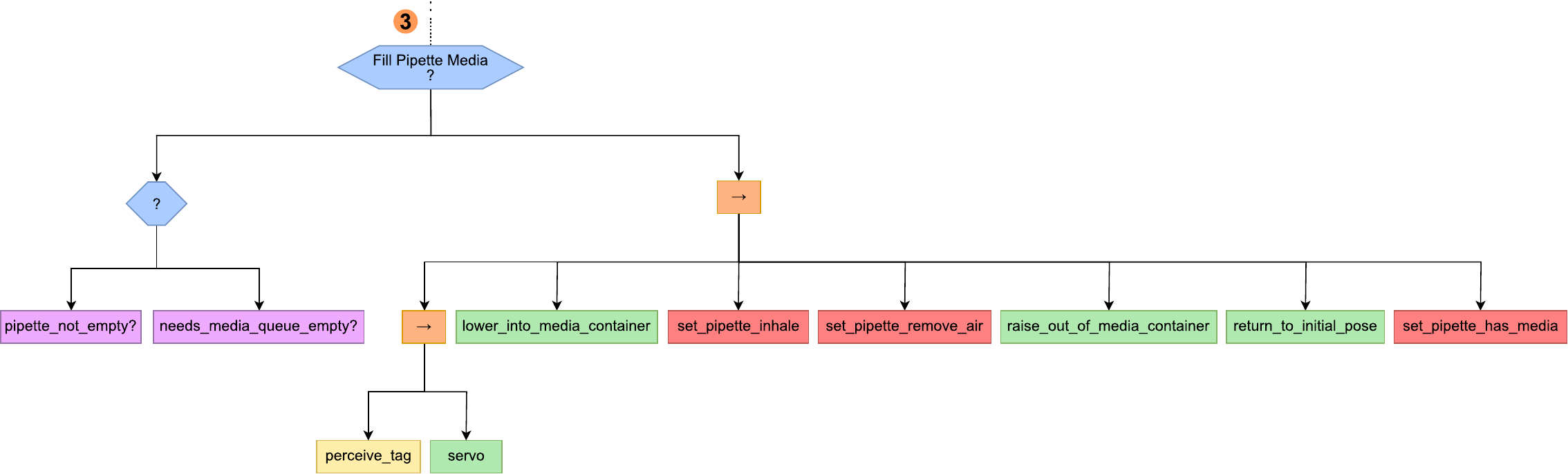}
    \caption{Behavior tree for \texttt{Fill Pipette - Media}. This sub-tree begins with a \texttt{Selector} that prevents redundant action by first checking whether the pipette is already filled or whether the queue of wells needing media is empty. If either condition is met, the tree returns \texttt{SUCCESS} without proceeding. Otherwise, the tree continues through a \texttt{Sequence} that executes the media-filling procedure. The robot first perceives the location of the media container and aligns itself using the \texttt{Servo} behavior. It then lowers into the container, inhales the liquid, performs an air removal routine, and raises the pipette back out. The pipette actuator length is controlled via \texttt{dynamic\_reconfigure}, so the \texttt{SetExperimentState} behavior is used to control the pipette as well. Lastly, the experiment state is updated to reflect that the pipette now contains media.}
    \label{fig:btree-fill-pipette-media}
\end{figure}

\begin{figure}[h]
    \centering
    \includegraphics[width=\linewidth]{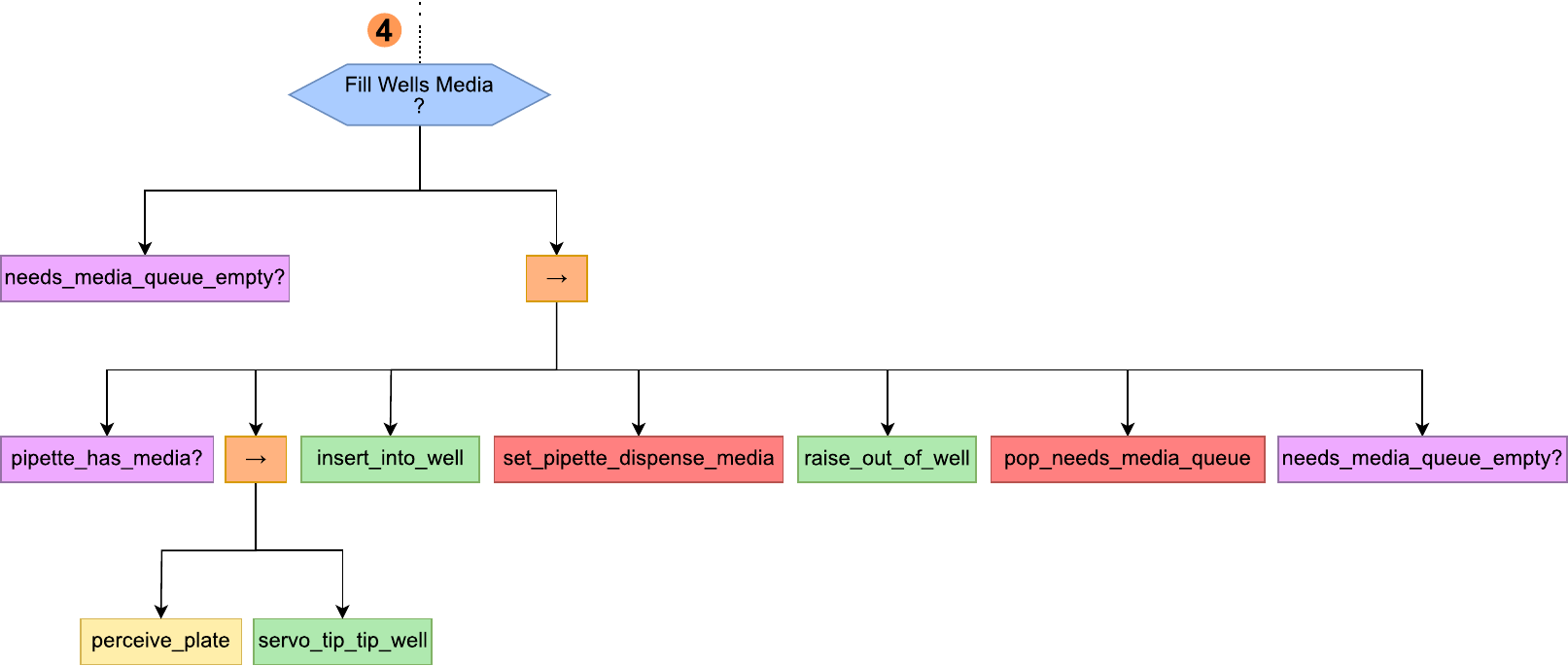}
    \caption{Behavior tree for \texttt{Fill Wells - Media}. This sub-tree handles the task of dispensing media into wells that require it. The root \texttt{Selector} first checks if the queue of wells needing media is empty. If not, it proceeds through a \texttt{Sequence} that performs the full dispensing routine. It begins by verifying that the pipette is filled with media. The robot then perceives the well plate and servo-aligns to the appropriate well before continuing. Once aligned, the robot inserts the pipette tip into the target well, dispenses media, and then raises the pipette back out. It then removes the serviced well from the queue and re-checks whether additional wells remain to be filled; the final \texttt{CheckExperimentState} in the sequence will fail if more wells require media, and the first \texttt{CheckExperimentState} ensures the pipette has not been emptied.}
    \label{fig:btree-fill-wells-media}
\end{figure}

\begin{figure}[h]
    \centering
    \includegraphics[width=\linewidth]{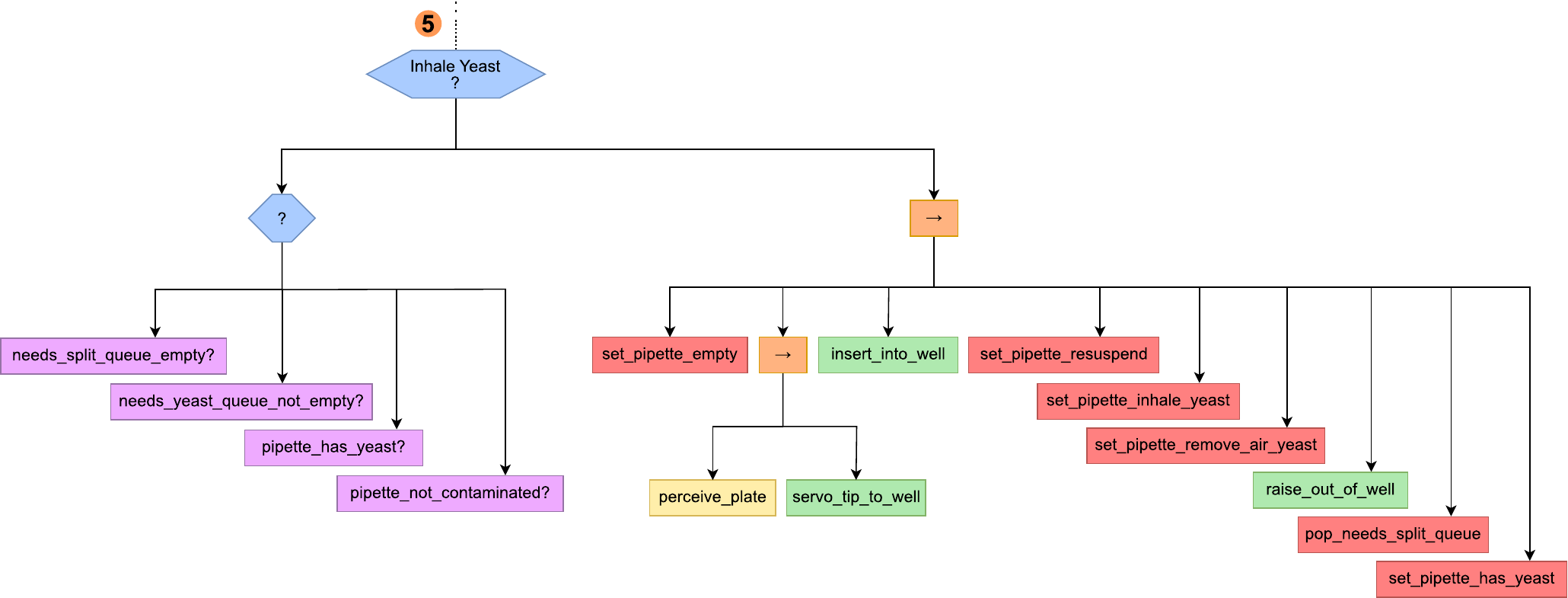}
    \caption{Behavior tree for \texttt{Inhale Yeast}. This sub-tree controls the process of resuspending and aspirating yeast from saturated wells. A series of preconditions are first checked, including whether the split queue is empty, the yeast queue is populated, the pipette already contains yeast, and whether the pipette is contaminated. If any of these checks fail, the behavior is aborted early. Otherwise, the robot proceeds through a \texttt{Sequence} to perform the aspiration. This begins with setting the pipette actuator to its maximum extension, followed by perceiving the well plate and servoing to the correct well. The robot then inserts the pipette tip into the well, resuspends the contents, inhales the yeast, removes excess air, and raises out of the well. Finally, it updates the experiment state by removing the well from the split queue and marking the pipette as containing yeast.}
    \label{fig:btree-inhale-yeast}
\end{figure}

\begin{figure}[h]
    \centering
    \includegraphics[width=\linewidth]{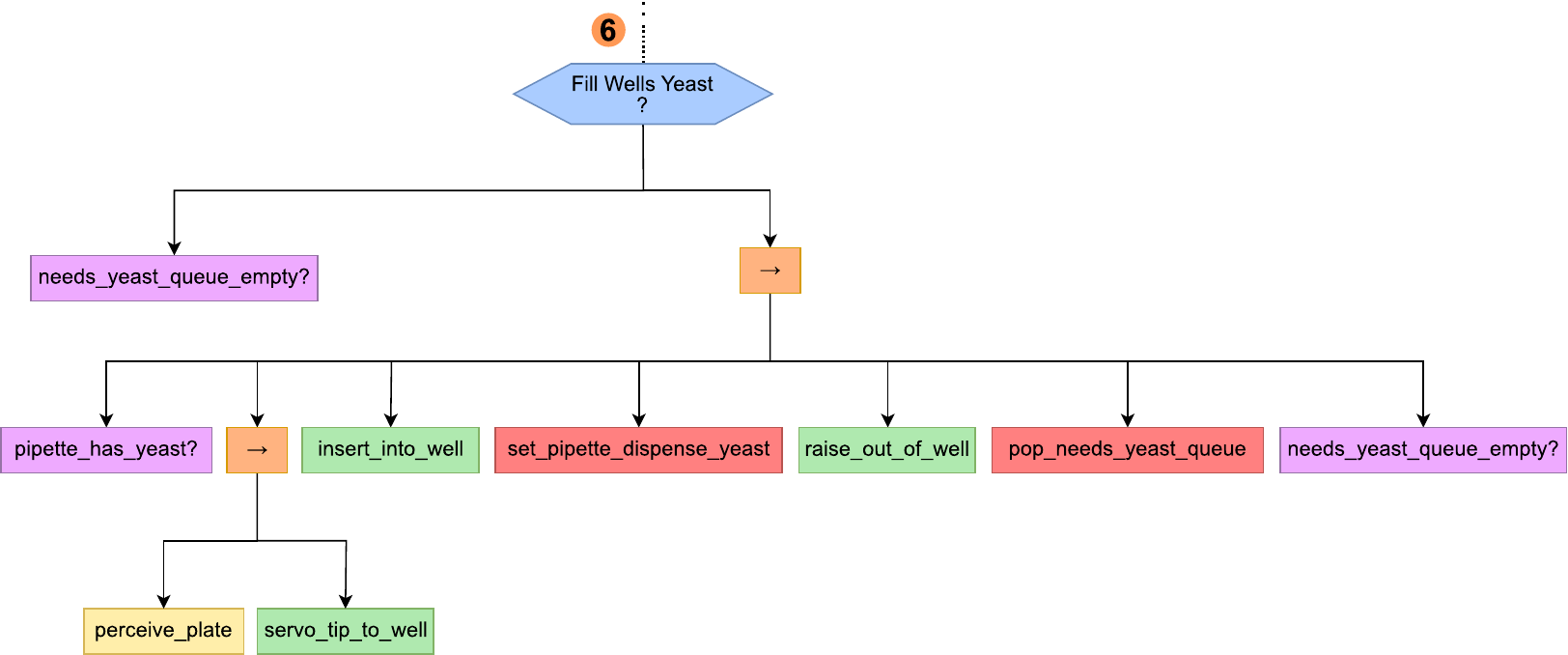}
    \caption{Behavior tree for \texttt{Fill Wells - Yeast}. This sub-tree begins by checking whether the queue of wells that require yeast is empty. If not, it proceeds through a \texttt{Sequence} that verifies the pipette is filled with yeast, perceives the well plate, and servo-aligns to the correct location. The pipette is then inserted into the well, yeast is dispensed, and the pipette is retracted. The completed well is removed from the queue, and the tree concludes by rechecking whether any wells remain.}
    \label{fig:btree-fill-wells-yeast}
\end{figure}

\begin{figure}[h]
    \centering
    \includegraphics[width=\linewidth]{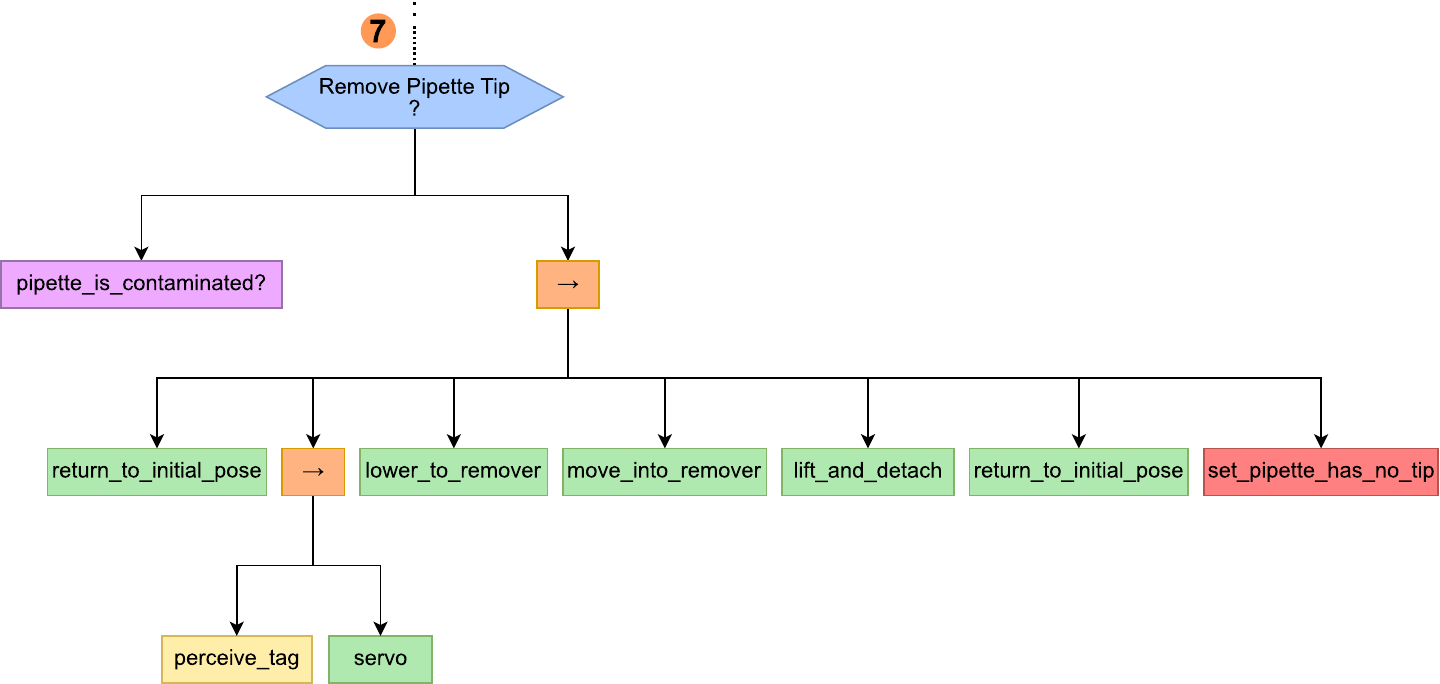}
    \caption{Behavior tree for \texttt{Remove Pipette Tip}. This sub-tree begins with a condition check on whether the pipette is contaminated, which is inverted, causing it to fail if true. If so, execution proceeds through a \texttt{Sequence} that perceives the tip removal stand and servo-aligns to it. The robot then lowers the pipette toward the remover, moves into position, and performs a lift-and-detach motion to release the tip. Afterward, it updates the experiment state to reflect that the pipette is no longer equipped with a tip.}
    \label{fig:btree-remove-tip}
    
\end{figure}

\begin{figure}[htbp]
    \centering
    \begin{minipage}[t]{0.48\linewidth}
        \centering
        \includegraphics[height=8cm]{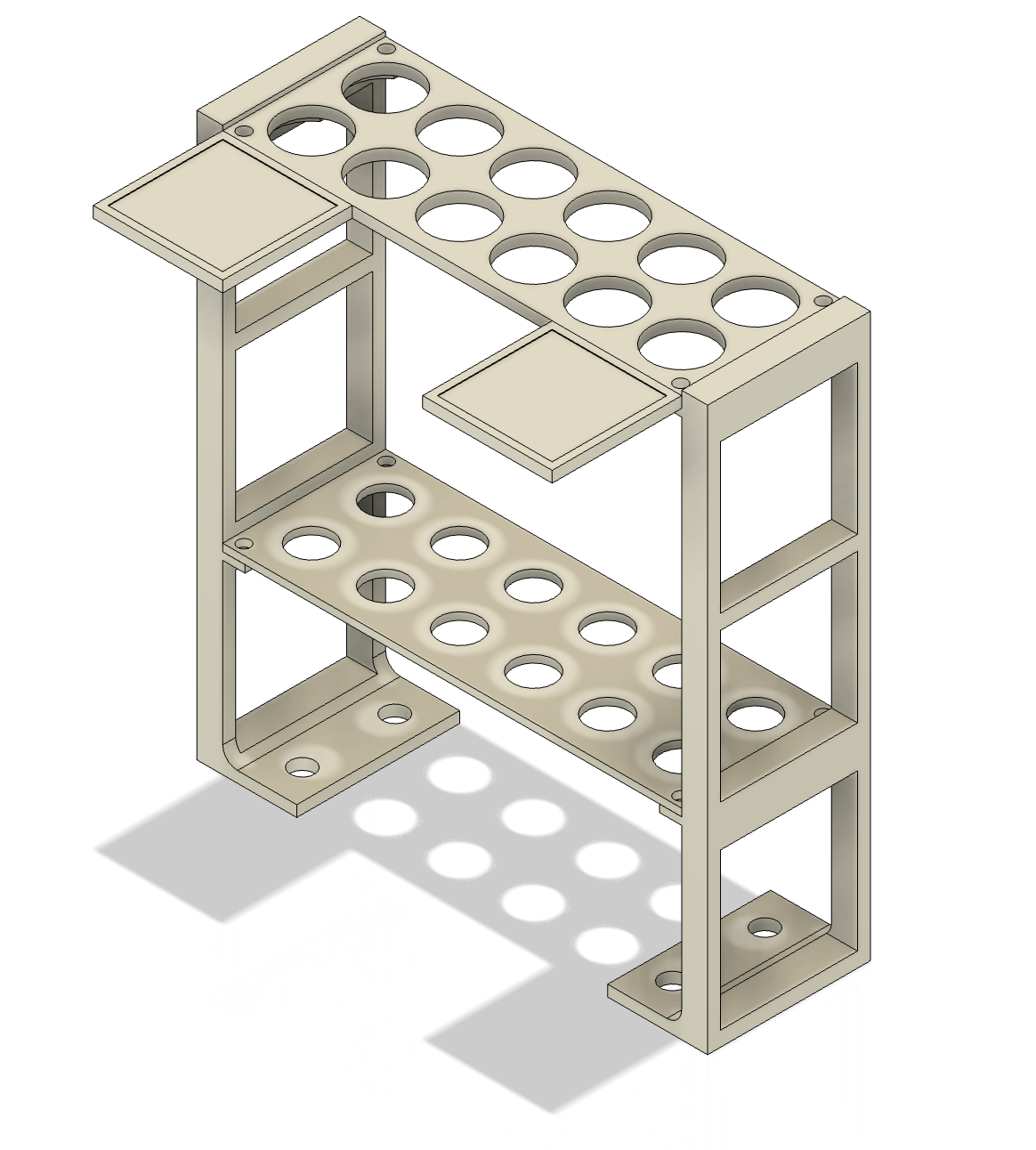}
        \caption{CAD Model of the pipette tip rack.}
        \label{fig:tiprack}
    \end{minipage}%
    \hfill
    \begin{minipage}[t]{0.48\linewidth}
        \centering
        \includegraphics[height=8cm]{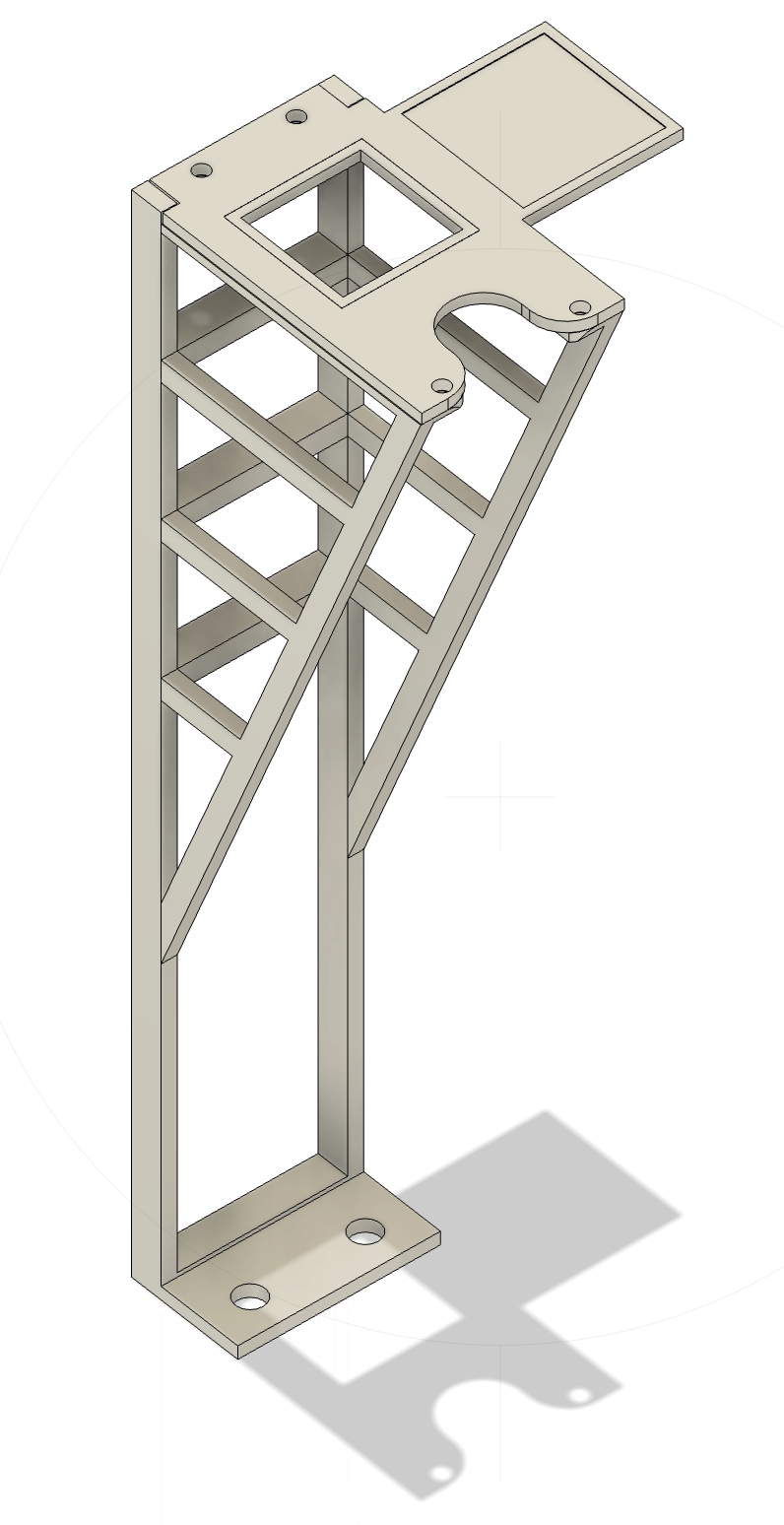}
        \caption{CAD Model of the pipette tip remover.}
        \label{fig:tipremover}
    \end{minipage}
\end{figure}

\begin{figure}[h]
    \centering
    \includegraphics[width=0.6\linewidth]{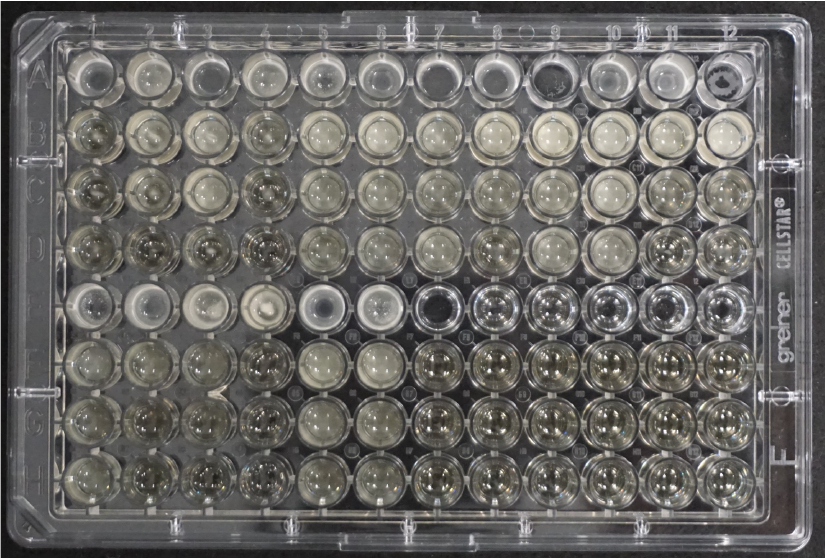}
    \caption{A photo of the well plate immediately after concluding the yeast cell culture experiment. Rows A and E contain the voided wells. Rows B-D, columns 1-6 contain the newly split wells of the 50 million cells/mL group. Rows B-D, columns 7-12 contain the newly split wells of the 30 million cells/mL group. Rows F-H, columns 1-6 contain the newly split wells of the 10 million cells/mL group. Rows F-H, columns 7-12 contain the newly split wells of the blank group.}
    \label{fig:final-well-plate}
\end{figure}

\begin{figure}[htbp]
    \centering
    \begin{minipage}{0.48\linewidth}
        \centering
        \includegraphics[width=\linewidth]{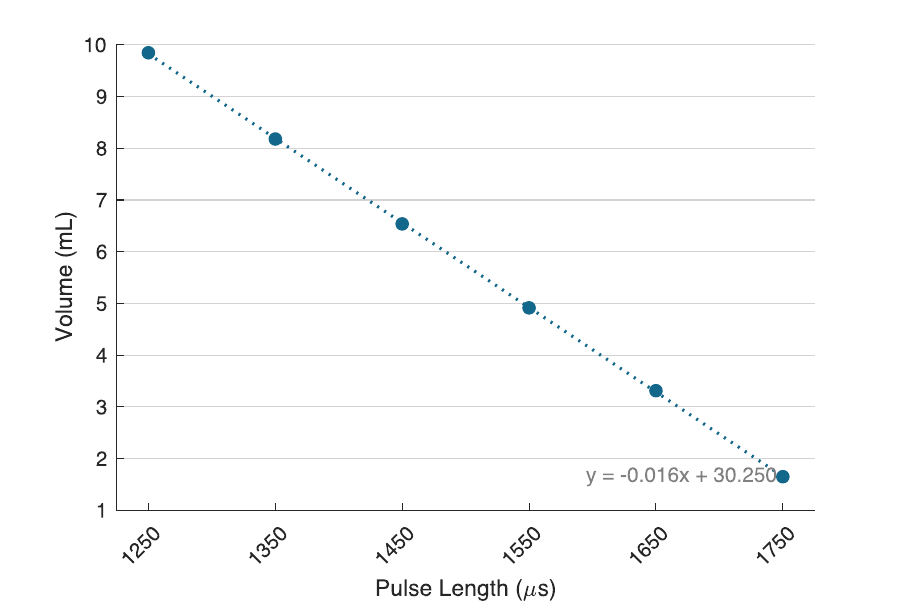}
        \caption{Pipette calibration plot for six pulse lengths between 1250 and 1750 microseconds. Error bars are included but not visible.}
        \label{fig:calib-all}
    \end{minipage}%
    \hfill
    \begin{minipage}{0.48\linewidth}
        \centering
        \includegraphics[width=\linewidth]{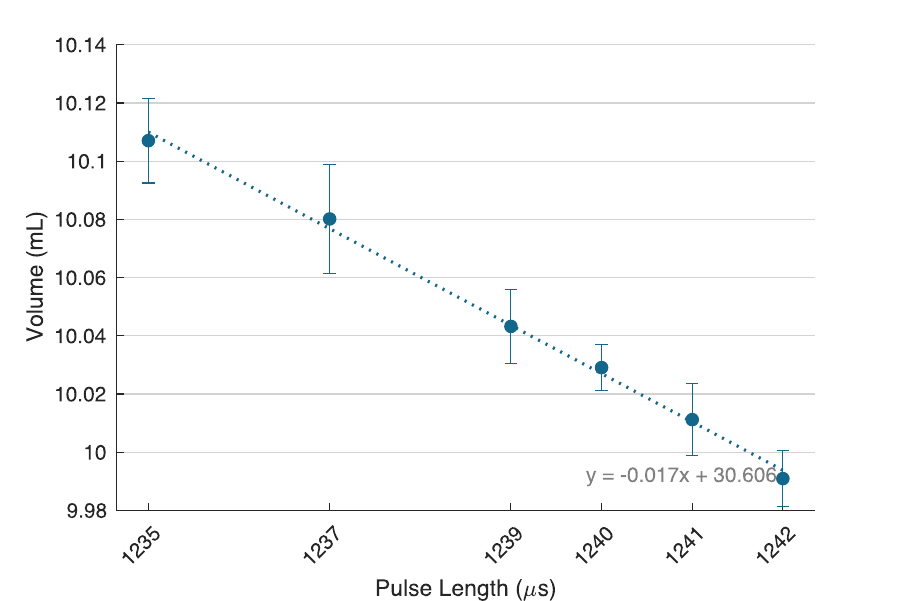}
        \caption{Pipette calibration plot for six pulse lengths between 1235 and 1242 microseconds, calibrating the pipette for 10 mL expulsion.}
        \label{fig:calib-10}
    \end{minipage}
\end{figure}

\begin{figure}[htbp]
    \centering
    \begin{minipage}{0.48\linewidth}
        \centering
        \includegraphics[width=\linewidth]{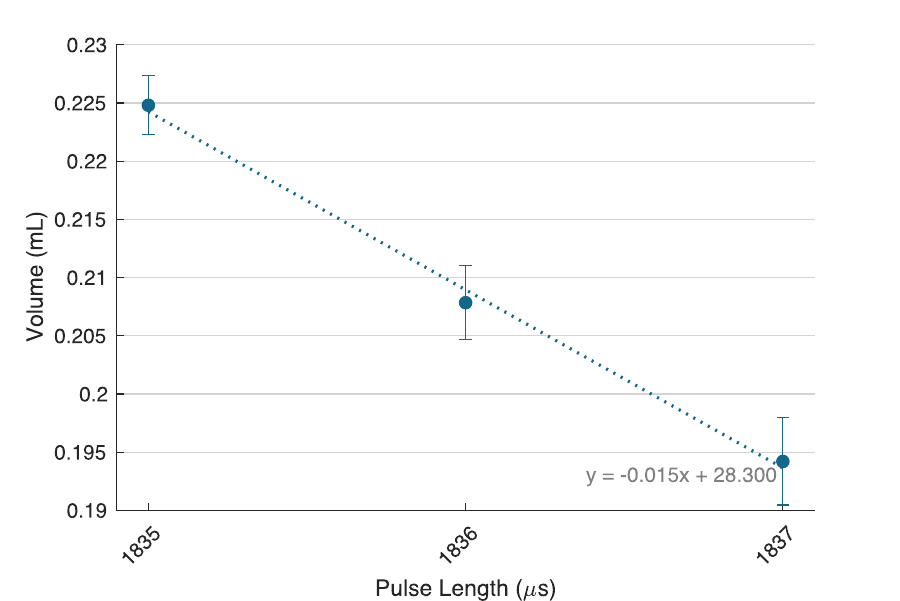}
        \caption{Pipette calibration plot for three pulse lengths between 1835 and 1837 microseconds, calibrating the pipette for 0.2 mL expulsion.}
        \label{fig:calib-02}
    \end{minipage}%
    \hfill
    \begin{minipage}{0.48\linewidth}
        \centering
        \includegraphics[width=\linewidth]{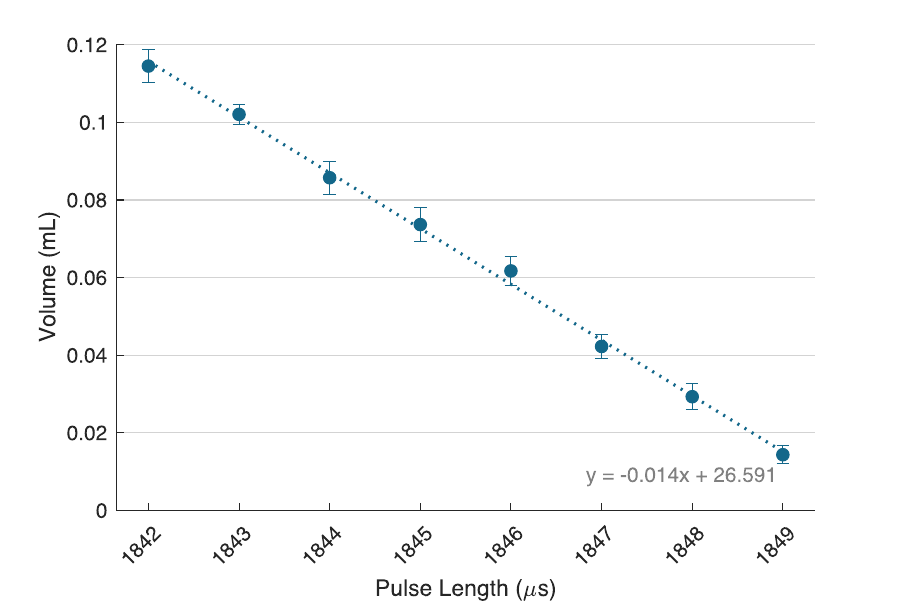}
        \caption{Pipette calibration plot for eight pulse lengths between 1842 and 1849 microseconds, representing the limit of the Digital Pipette v2 for small volumes. }
        \label{fig:calib-01}
    \end{minipage}
\end{figure}

Execution begins at the root and propagates downward through the tree in ticks, which are periodic updates that determine which actions should be run. Control nodes manage the order and logic of their child nodes, while leaf nodes perform specific behaviors. This structure enables the robot to respond dynamically to changes in the environment, retry failed actions, or prioritize critical operations without hard-coding rigid state transitions. Compared to finite state machines, behavior trees offer greater flexibility, reusability, and scalability. \textbf{Figures \ref{fig:btree-main} - \ref{fig:btree-remove-tip}} represent the behavior trees used for RoboCulture.

Each node in a behavior tree must return one of three statuses when ticked:
\begin{itemize}
    \item \verb|SUCCESS|: The node has completed its task successfully
    \item \verb|FAILURE|: The node was unable to complete its task
    \item \verb|RUNNING|: The node is still in progress and requires further ticks
\end{itemize}

These return statuses propagate upward and determine how control nodes (e.g., Sequence or Selector) behave. This mechanism allows behavior trees to remain reactive, handling task interruptions and conditional logic gracefully.

The control nodes govern the execution of their children. Two of the most commonly used control flow types are the Sequence and the Selector.

\paragraph{\textcolor{BurntOrange}{Sequence}}
The Sequence node (“AND” node) ticks its children from left to right. It returns \verb|SUCCESS| only if all child nodes return \verb|SUCCESS|, \verb|FAILURE| immediately when any child returns \verb|FAILURE|, \verb|RUNNING| if a child returns \verb|RUNNING|, pausing execution until the next tick.

This structure is useful for defining a series of dependent steps, such as “grasp pipette,” “attach pipette tip,” “pipette media”, where each step must succeed for the overall task to complete. 
\\
\paragraph{\textcolor{RoyalBlue}{Selector}}
The Selector node (“OR” node) also ticks its children from left to right, but returns \verb|SUCCESS| as soon as any child returns \verb|SUCCESS|, \verb|FAILURE| only if all children return \verb|FAILURE|, \verb|RUNNING| if a child returns \verb|RUNNING|, pausing execution until the next tick.

This node is ideal for fallback strategies or priorities. For example, the \verb|Grasp Pipette| behavior tree in \textbf{Figure \ref{fig:btree-grasp-pipette}} is defined using a Selector. Its left child checks if the \verb|holding_pipette| state is \verb|True|; if so, it returns \verb|SUCCESS| and the \verb|Grasp Pipette| tree immediately returns \verb|SUCCESS|. Otherwise, the left child will return \verb|FAILURE|, and will proceed to execute the right child, which defines a Sequence to grasp the pipette. 
\\
\newline
The following behaviors were built for RoboCulture:

\paragraph{\textcolor{red}{SetExperimentState}}

RoboCulture uses the \verb|dynamic_reconfigure| ROS package to manage parameters which are useful for the execution of the experiment. The package allows for the reconfiguration of parameters during execution, and also provides a graphical user interface. More information on the tunable parameters is provided in \textbf{Section \ref{sec:rqt}}.
The \verb|SetExperimentState| behavior is used to programatically set, increment, or decrement one of the parameters stored in the dynamic reconfigure server. It always returns \verb|SUCCESS|. 
\\
\paragraph{\textcolor{Thistle}{CheckExperimentState}}

The \verb|CheckExperimentState| behavior evaluates the value of a runtime parameter exposed through the \verb|dynamic_reconfigure| server to control conditional branching in the behavior tree. The behavior evaluates the current value of a specified parameter against a desired target value using a comparison operator (e.g., equals, greater than, less than). If the condition holds, the behavior returns \verb|SUCCESS|; otherwise, it returns \verb|FAILURE|.

This node is often used to guard access to downstream actions, such as checking whether a pipette is currently held before attempting to grasp one.
\\
\paragraph{\textcolor{Goldenrod}{Perceive}}

The \verb|Perceive| behavior initiates a perception request via a ROS action server to the \verb|perception_node|, instructing it to search for a specified target in the scene. The node accepts as parameters the target type and an identifier (e.g., AprilTag ID or well number).

The behavior returns \verb|RUNNING| while the perception node is actively processing the goal. If perception completes successfully and the \verb|perception_node| begins publishing error vectors, the \verb|Perceive| behavior returns \verb|SUCCESS|. If perception fails (e.g., the target is not detected), it returns \verb|FAILURE|.
\\
\paragraph{\textcolor{ForestGreen}{Servo}}

The \verb|Servo| behavior is used to command the robot’s end-effector toward a target pose, either through perception-guided visual feedback or predefined Cartesian displacements. In both cases, the behavior communicates with the robot’s \verb|control_node| by sending a goal via a ROS action server, which handles the underlying motion execution. When operating in visual servoing mode, \verb|Servo| minimizes a perception error vector provided by the perception node to align a target with a detected object such as an AprilTag or well. Alternatively, it can move the end-effector by a specified offset or an absolute position in the robot’s base frame, which are useful for open-loop motions like insertion, retraction, or coarse positioning.

\verb|Servo| returns \verb|RUNNING| while the robot is in motion, \verb|SUCCESS| when the goal is achieved, and \verb|FAILURE| if the target becomes undetectable or the motion is aborted. In the case of failure, the behavior cancels any active goals and halts the robot, allowing the tree to fallback and re-initialize its perception target.

\FloatBarrier

\FloatBarrier
\subsection{CAD Models}
\label{sec:cad-models}

\FloatBarrier
\subsection{Object Position Estimation Error Due to Hand-Eye Calibration Error}
\label{sec:perception-error}
The upper bound on the object‑position error,
\[
\|\delta^{\mathcal{I}}\bm p_{\mathcal O}\|
\;\le\;
\|\bm R_\epsilon-\bm I\|\,\bigl\|^{\mathcal{C}}\bm p_{\mathcal O}\bigr\|
+\|\bm p_\epsilon\|,
\]
shows that \emph{the farther the object is from the camera (\(\|^{\mathcal{C}}\bm p_{\mathcal O}\|\) is larger), the larger the world‑frame position error produced by extrinsic‑calibration inaccuracies}.  In other words, rotational mis-alignment contributes a distance‑scaled error term that grows linearly with object–camera separation, while the translational mis-registration \(\|\bm p_\epsilon\|\) adds a constant offset.

${\mathcal{I}}$ is the inertial or the world frame, ${\mathcal{C}}$ is the camera frame, and ${\mathcal{T}}$ is the robot tool or end-effector frame.
We assume that the errors of object pose estimation in the camera frame and the robot tool pose estimation using the robot encoders are negligible compared to the hand-eye calibration error.
We can write down:
\begin{equation}
\begin{array}{l}
^{\mathcal{I}}\tilde{\bm{T}}_{\mathcal{O}} = ~^{\mathcal{I}}\tilde{\bm{T}}_{\mathcal{T}}~ ^{\mathcal{T}}\tilde{\bm{T}}_{\mathcal{C}}~ ^{\mathcal{C}}\tilde{\bm{T}}_{\mathcal{O}}
 = \begin{pmatrix}
  ^{\mathcal{I}}\bm{R}_{\mathcal{O}} & ^{\mathcal{I}}{\bm{p}}_{\mathcal{O}} \\
  \bm{0} & 1 
 \end{pmatrix} = 
 \\ 
 \begin{pmatrix}
  ^{\mathcal{I}}\bm{R}_{\mathcal{T}} & ^{\mathcal{I}}{\bm{p}}_{\mathcal{T}} \\
  \bm{0} & 1 
 \end{pmatrix}
  \begin{pmatrix}
  ^{\mathcal{T}}\bm{R}_{\mathcal{C}} ~ \bm{R}_{\epsilon} & ^{\mathcal{T}}{\bm{p}}_{\mathcal{C}} + \bm{p}_{\epsilon}\\
  \bm{0} & 1 
 \end{pmatrix}
\begin{pmatrix}
  ^{\mathcal{C}}\bm{R}_{\mathcal{O}} & ^{\mathcal{C}}{\bm{p}}_{\mathcal{O}} \\
  \bm{0} & 1 
\end{pmatrix},
\end{array}
\end{equation}
where $\bm{R}_{\epsilon}$ and $\bm{p}_{\epsilon}$ result from the camera calibration error. In the case of no error, $\bm{R}_{\epsilon} = \bm{I}$ and $\bm{p}_{\epsilon}=\bm{0}$. From this equation, we can compute $ \delta ^{\mathcal{I}}\bm{p}_{\mathcal{O}}$ as:
\begin{align}
\delta ^{\mathcal{I}}\bm{p}_{\mathcal{O}} 
&= \,^{\mathcal{I}}\tilde{\bm{p}}_{\mathcal{O}} - \,^{\mathcal{I}}\bm{p}_{\mathcal{O}} \nonumber \\
&= \,^{\mathcal{I}}\bm{R}_{\mathcal{T}} \,^{\mathcal{T}}\bm{R}_{\mathcal{C}} (\bm{R}_{\epsilon} - \bm{I}) \,^{\mathcal{C}}\bm{p}_{\mathcal{O}} 
    + \,^{\mathcal{I}}\bm{R}_{\mathcal{T}} \,\bm{p}_{\epsilon}
\end{align}
Using the triangle inequality for vectors and matrix norm submultiplicative lemmas, we can deduce:
\begin{align}
\| \delta ^{\mathcal{I}}\bm{p}_{\mathcal{O}} \|
&= \left\| ^{\mathcal{I}}\bm{R}_{\mathcal{T}} \, ^{\mathcal{T}}\bm{R}_{\mathcal{C}} \, ( \bm{R}_{\epsilon} - \bm{I}) \, ^{\mathcal{C}}\bm{p}_{\mathcal{O}} 
     + ^{\mathcal{I}}\bm{R}_{\mathcal{T}} \, \bm{p}_{\epsilon} \right\| \nonumber \\
&\leq \|^{\mathcal{I}}\bm{R}_{\mathcal{T}}\| \, \|^{\mathcal{T}}\bm{R}_{\mathcal{C}}\| \, \| \bm{R}_{\epsilon} - \bm{I} \| \, \|^{\mathcal{C}}\bm{p}_{\mathcal{O}}\| 
     + \|^{\mathcal{I}}\bm{R}_{\mathcal{T}}\| \, \|\bm{p}_{\epsilon}\| \nonumber \\
&\leq \| \bm{R}_{\epsilon} - \bm{I} \| \, \|^{\mathcal{C}}\bm{p}_{\mathcal{O}}\| + \| \bm{p}_{\epsilon} \|.
\end{align}
This inequality therefore shows that, when the rotational portion of the calibration error is non‑zero ($\bm{R}_{\epsilon}\neq\bm{I}$), the upper bound on the object‑position estimation error \emph{increases linearly with the camera–object distance}. That is, a larger $\|^{\mathcal{C}}\bm{p}_{\mathcal{O}}\|$ yields a larger error.

\newpage

\FloatBarrier
\subsection{Final Well Plate}
\label{sec:final-plate}

\FloatBarrier
\subsection{Pipette Calibration Plots}
\label{sec:pipette-calibration}

\FloatBarrier
\subsection{RQT Graphical User Interface}
\label{sec:rqt}

\begin{figure}[htbp]
    \centering
    \begin{minipage}{0.48\linewidth}
        \centering
        \includegraphics[width=\linewidth]{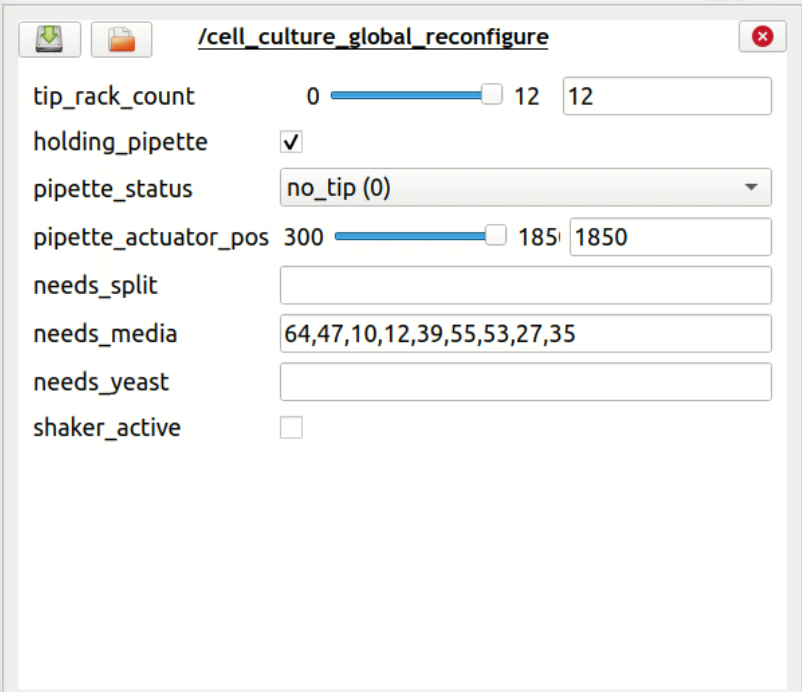}
    \end{minipage}%
    \hfill
    \begin{minipage}{0.48\linewidth}
        \centering
        \includegraphics[width=\linewidth]{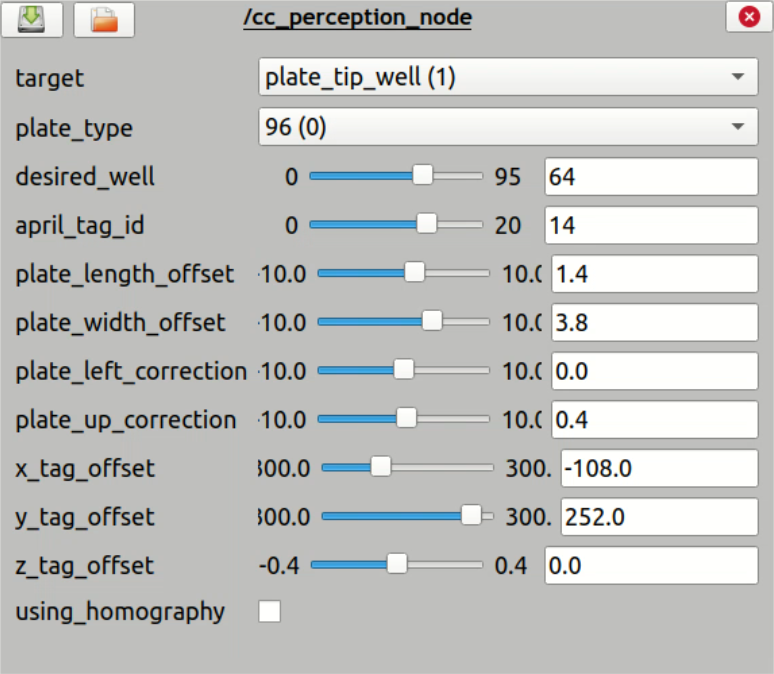}
    \end{minipage}
    \caption{\texttt{dynamic\_reconfigure} RQT interface used to visualize parameters during experiment execution, and for debugging purposes.}
    \label{fig:rqt}
\end{figure}

The \verb|dynamic_reconfigure| ROS package is used in conjunction with an rqt-based graphical user interface to provide access to the variables used to manage the experiment state, which can be tuned in real-time. These fields can be updated manually or programmatically by the behavior tree to reflect the current needs of the workflow. \textbf{Figure \ref{fig:rqt}} shows the user interface. The left panel displays parameters related to high-level experiment state. A brief description of each parameter is given below.
\begin{itemize}
    \item \texttt{tip\_rack\_count}: Ranges from 0-12, represents the number of pipette tips which remain in the tip rack
    \item \texttt{holding\_pipette}: Binary value reflecting whether the robot is holding the pipette
    \item \texttt{pipette\_status}: Takes on a series of values based on the current status of the pipette. (0): the pipette has no tip attached, (1): the pipette tip is attached and it is empty, (2): the pipette has media, (3): the pipette has yeast, (4): the pipette tip is empty and contaminated. 
    \item \texttt{pipette\_actuator\_pos}: Ranges from 1300 to 1850, and represents the length of the pulse in microseconds to send to the pipette actuator to control its length. 1300 corresponds to maximally retracted (inhale), and 1850 corresponds to maximally extended (exhale). Changing this value will immediately take effect on the Digital Pipette.
    \item \texttt{needs\_split}: The queue of wells that need to be split.
    \item \texttt{needs\_media}: The queue of wells that need media.
    \item \texttt{needs\_yeast}: The queue of wells that need yeast.
    \item \texttt{shaker\_active}: A binary value representing the status of the orbital shaker platform. Changing this value will immediately stop/start the shaker.
\end{itemize}

The right panel displays parameters related to perception. A brief description of each parameter is given below.
\begin{itemize}
    \item \texttt{target}: Takes on a value based on the current mode of perception. (0): No perception target, (1): \texttt{plate\_tip\_well}, align the pipette tip with the desired well, (2): \texttt{plate\_cam\_well}, align the center of the image to the desired well, (3): \texttt{plate\_cam\_center}, align the center of the image to the center of the well plate, (4): \texttt{april\_tag}, align the center of the camera to the center of the AprilTag. 
    \item \texttt{plate\_type}: Takes on a value to identify the size of the well plate. (0): 96-well plate, (1): 24-well plate, (2): 6-well plate.
    \item \texttt{desired\_well}: Ranges from 0 to 95, used to select the desired well.
    \item \texttt{april\_tag\_id}: Ranges from 0 to 20, used to select the desired AprilTag id.
    \item \texttt{plate\_length\_offset}: Used to offset the length of the template well plate.
    \item \texttt{plate\_width\_offset}: Used to offset the width of the template well plate.
    \item \texttt{plate\_left\_correction}: Used to offset the starting left value of the template well plate.
    \item \texttt{plate\_up\_correction}: Used to offset the starting top value of the template well plate.
    \item \texttt{x\_tag\_offset}: Used to offset the x coordinate of the detected April Tag (px).
    \item \texttt{y\_tag\_offset}: Used to offset the y coordinate of the detected April Tag (px).
    \item \texttt{z\_tag\_offset}: Used to offset the z coordinate of the detected April Tag (px).
    \item \texttt{using\_homography}: Binary value representing whether to use homography to correct camera perspective shift.
\end{itemize}

\FloatBarrier
\newpage
\subsection*{Supplementary Videos}
\noindent
\textbf{Supplementary Video 1.} The pipette tip insertion experiment (50x speed). Repeated insertions of the pipette tip into the wells of a 96-well plate in a random order are performed to evaluate the reliability of the system. The test was repeated at six unique well plate positions, with 96 insertions per well plate position. 
\\

\noindent
\textbf{Supplementary Video 2.} Close up of the pipette tip attachment system (1x speed). There is initially misalignment between the end of the pipette body and the opening of the new pipette tip. The robot presses the pipette body onto the edge of the new tip as it is guided in a spiral motion. The force perceived from the environment is monitored, and the robot inserts the pipette into the new tip once the force drops, indicating the correct alignment was found.
\\

\noindent
\textbf{Supplementary Video 3.} Close up of the pipette tip detachment system (1x speed). The robot guides the pipette such that a 3D printed hook affixed to the work table pulls down on the pipette tip, detaching it from the body. The contaminated pipette tip falls into a biological waste bin. 
\\

\noindent
\textbf{Supplementary Video 4.} The pipette tip attachment experiment (10x speed). Repeated pipette tip attachments and detachments are performed to evaluate the reliability of the system. 
\\

\noindent
\textbf{Supplementary Video 5.} Demonstration of the well plate perception system (1x speed). The robot end effector is moved around manually, and the output from the perception pipeline is visualized. The green dots represent the perceived centers of the 96 wells. 
\\

\noindent
\textbf{Supplementary Video 6.} Demonstration of the visual servoing system (1x speed). While someone disturbs the position of the well plate, the robot reacts in real-time to guide the pipette tip in line with the desired well. 
\\

\noindent
\textbf{Supplementary Video 7.} The optical density monitoring procedure (2x speed). The robot pans across the well plate, taking images of the wells in batches. The lower right shows the output from the camera affixed to the robot end effector–the green circles represent the detected locations of the twelve wells which are next to be imaged.
\\

\noindent
\textbf{Supplementary Video 8.} Full run of the yeast cell culturing experiment. The four splitting processes are shown at 60x speed, and the idle and optical density monitoring processes are shown at 200x speed. The full duration of the experiment lasted over 15 hours.
\\

\noindent
\textbf{Supplementary Video 9.} Expansion of the 50 million/mL group (30x speed). After completing the optical density monitoring procedure, the robot initiates the well-splitting process. It first retrieves the pipette from its stand and attaches a fresh pipette tip. Growth media is then aspirated and dispensed into 18 empty wells. The robot proceeds by resuspending the first saturated well and distributing its contents into three of the pre-filled wells, followed by the disposal of the used pipette tip. For each of the remaining five saturated wells, a new pipette tip is attached, used, and discarded in sequence.
\\

\noindent
\textbf{Supplementary Video 10.} Side view of the expansion of the 50 million/mL group (20x speed).
\\

\noindent
\textbf{Supplementary Video 11.} Close up of the well splitting process (5x speed). First, the saturated well is resuspended to mitigate the yeast settling on the bottom of the plate. Then, 0.05 mL of the saturated liquid is dispensed into each of the three wells pre-filled with culture media. 
\\

\noindent
\textbf{Supplementary Video 12.} View from the camera affixed to the robot during the well passaging procedure. The left panel displays the raw image, while the right panel shows the annotated image used for visualization. A blue mask highlights the pipette, representing the segmentation output from the perception pipeline. Green circles indicate the detected positions of the 96 wells. The annotated video stops when well plate perception is disengaed.
\\

\noindent
\textbf{Supplementary Video 13.} A montage showing the saturation of the yeast in the well plate. Each frame represents about 7 minutes of growth.

\end{document}